\documentclass[times,sort&compress,review,10pt,number,3p]{elsarticle}

\usepackage{amssymb}
\usepackage{xspace}
\usepackage{float}
\usepackage{lineno}
\usepackage{booktabs}
\usepackage{multirow}
\usepackage{bigstrut}

\usepackage[caption=false,font=footnotesize,labelfont=rm,textfont=rm]{subfig}

\usepackage{multirow}
\usepackage{graphicx}
\usepackage{booktabs}
\usepackage{array}
\usepackage{xcolor}
\usepackage{amsmath}
\usepackage{pifont}

\newcommand{\ie}{\textit{i.e.}\xspace}
\newcommand{\eg}{\textit{e.g.}\xspace}

\newcommand{\etc}{\textit{etc.}\xspace}

\usepackage{hyperref}
\usepackage[capitalize]{cleveref}
\crefname{section}{Sec.}{Secs.}
\crefname{table}{Table}{Tables}
\crefname{figure}{Fig.}{Figs.}
\journal{Pattern Recognition}

\makeatletter
\let\oldsubsection\subsection
\renewcommand\subsection{\@afterindenttrue\oldsubsection}
\renewcommand\subsection{\@startsection{subsection}{2}{\z@}%
  {-3.25ex\@plus -1ex \@minus -.2ex}%
  {1.5ex \@plus .2ex}%
  {\normalfont\normalsize}}

\let\oldsubsubsection\subsubsection
\renewcommand\subsubsection{\@afterindenttrue\oldsubsubsection}
\renewcommand\subsubsection{\@startsection{subsubsection}{3}{\z@}%
  {-3.25ex\@plus -1ex \@minus -.2ex}%
  {1.5ex \@plus .2ex}%
  {\normalfont\normalsize}}
\makeatother

\DeclareSubrefFormat{parens}{#1(#2)}
\begin{document}

\begin{frontmatter}

     %% Title, authors and addresses

     %% use the tnoteref command within \title for footnotes;
     %% use the tnotetext command for theassociated footnote;
     %% use the fnref command within \author or \affiliation for footnotes;
     %% use the fntext command for theassociated footnote;
     %% use the corref command within \author for corresponding author footnotes;
     %% use the cortext command for theassociated footnote;
     %% use the ead command for the email address,
     %% and the form \ead[url] for the home page:
     %% \title{Title\tnoteref{label1}}
     %% \tnotetext[label1]{}
     %% \author{Name\corref{cor1}\fnref{label2}}
     %% \ead{email address}
     %% \ead[url]{home page}
     %% \fntext[label2]{}
     %% \cortext[cor1]{}
     %% \affiliation{organization={},
     %%             addressline={},
     %%             city={},
     %%             postcode={},
     %%             state={},
     %%             country={}}
     %% \fntext[label3]{}

     \title{CLIP-driven rain perception: Adaptive deraining with pattern-aware network routing and mask-guided cross-attention}

     %% use optional labels to link authors explicitly to addresses:
     %% \author[label1,label2]{}
     %% \affiliation[label1]{organization={},
     %%             addressline={},
     %%             city={},
     %%             postcode={},
     %%             state={},
     %%             country={}}
     %%
     %% \affiliation[label2]{organization={},
     %%             addressline={},
     %%             city={},
     %%             postcode={},
     %%             state={},
     %%             country={}}

     \author[1]{Cong GUAN}
     \address[1]{Graduate School of Information, Production and Systems, Waseda University, Fukuoka, 808-0135, Japan}
     \ead{guancong@fuji.waseda.jp}

     \author[1]{Osamu YOSHIE\corref{cor1}}
     \ead{yoshie@waseda.jp}

     \cortext[cor1]{Corresponding author}

     %% Abstract
     \begin{abstract}
          %% Text of abstract
          Existing deraining models process all rainy images within a single network. However, different rain patterns have significant variations, which makes it challenging for a single network to handle diverse types of raindrops and streaks. To address this limitation, we propose a novel CLIP-driven rain perception network (CLIP-RPN) that leverages CLIP to automatically perceive rain patterns by computing visual-language matching scores and adaptively routing to sub-networks to handle different rain patterns, such as varying raindrop densities, streak orientations, and rainfall intensity. CLIP-RPN establishes semantic-aware rain pattern recognition through CLIP's cross-modal visual-language alignment capabilities, enabling automatic identification of precipitation characteristics across different rain scenarios. This rain pattern awareness drives an adaptive subnetwork routing mechanism where specialized processing branches are dynamically activated based on the detected rain type, significantly enhancing the model's capacity to handle diverse rainfall conditions. Furthermore, within sub-networks of CLIP-RPN, we introduce a mask-guided cross-attention mechanism (MGCA) that predicts precise rain masks at multi-scale to facilitate contextual interactions between rainy regions and clean background areas by cross-attention. We also introduces a dynamic loss scheduling mechanism (DLS) to adaptively adjust the gradients for the optimization process of CLIP-RPN. Compared with the commonly used $l_1$ or $l_2$ loss, DLS is more compatible with the inherent dynamics of the network training process, thus achieving enhanced outcomes. Our method achieves state-of-the-art performance across multiple datasets, particularly excelling in complex mixed datasets. 
     \end{abstract}

     % %%Graphical abstract
     % \begin{graphicalabstract}
     % %\includegraphics{grabs}
     % \end{graphicalabstract}

     \begin{highlights}
          \item CLIP-driven rain perception enables adaptive routing for diverse rain patterns.
          \item Mask-guided cross-attention enhances feature interaction between rainy and non-rainy regions.
          \item Dynamic loss scheduling aligns with network optimization process, enhancing training effectiveness.
          \item State-of-the-art performance achieved across multiple datasets, excelling in complex mixed scenarios.
     \end{highlights}

     %% Keywords

\end{frontmatter}

%\linenumbers

\section{Introduction}\label{sec:intro}

As one of the most common weather conditions, rain substantially deteriorates the quality of images and video captured in outdoor environments~\cite{10945649,yang2020single,ZHANG2023109740}. Whether in surveillance systems, autonomous vehicles, or photography, rain streaks often obscure important visual details, hindering the performance of various vision-based applications. Image deraining is the process of removing rain from images, which is crucial for enhancing the quality of visual data~\cite{,su2023survey,Wang2022RainRemovalSurvey} and plays a vital role in ensuring the reliable performance of computer vision tasks, such as object detection, tracking, and recognition, \etc

Traditionally, deraining methods have been based on model-driven techniques, such as filter-based and prior-based approaches~\cite{Wang2022RainRemovalSurvey}. Filter-based methods employ physical models to remove rain, while prior-based techniques use optimization frameworks with assumptions such as sparsity~\cite{Gu_2017_ICCV} or low-rank structures~\cite{Li_2016_CVPR} to estimate clean images. With the advancement of machine learning, especially deep learning (DL), data-driven methods have taken center stage, particularly those leveraging convolutional neural networks~\cite{AlexNet} (CNNs) and generative adversarial networks~\cite{GANs} (GANs). These methods have achieved impressive results in removing rain streaks by learning complex mappings from rainy to clean images. More recently, the emergence of Transformer-based~\cite{Attention,ViT} models and diffusion-based~\cite{diffusion} techniques has further enhanced deraining performance, capturing long-range dependencies and non-local patterns that are crucial for accurately restoring images affected by diverse rain conditions. Despite these advances, existing deraining techniques still face notable challenges and limitations.

\textbf{Complex Rain Patterns}. A major limitation of current rain removal methods lies in their ability to handle complex rain patterns. Rain can exhibit significant variability in terms of intensity, direction, and streak morphology~\cite{ZHANG2023109740, yang2020single, Wang2022RainRemovalSurvey}, and current methods often struggle to generalize across such variations. For instance, conventional deep learning approaches often struggle with this variability due to their inherent spatial invariance and limited receptive fields, which may hinder their capacity to accurately model the diverse and irregular structures of rain streaks. Moreover, most existing deep models adopt a single-path architecture that processes all rain types within a unified framework. This design can be insufficient for capturing the full complexity of rain patterns, frequently resulting in artifacts or incomplete restoration in more challenging scenarios.

\textbf{Rain and Clear Region Interaction}. Another limitation lies in the interaction between rainy and non-rainy regions within the image. Rainy regions often contain obscured or noisy image content, and the primary challenge is to restore these areas while retaining the details in the non-rainy parts. Non-rainy regions, by contrast, tend to contain clearer, more detailed information that is crucial for accurate restoration. However, existing models often lack effective mechanisms for facilitating the interaction between these two complementary parts. As a result, these methods may fail to properly guide the restoration process, leading to suboptimal performance in cases where both rain and fine details need to be simultaneously addressed. Without a clear distinction between rainy and non-rainy areas and a mechanism for their interaction, the overall deraining effect can be compromised.

\textbf{Optimization Strategy}. Another important aspect to consider is the optimization strategy during model training. Traditional loss functions, such as $l_1$ loss, treat all pixels equally, which doesn't account for the natural distribution of information in an image or the natural progression in neural network optimization. In real-world images, low-frequency smooth areas typically dominate spatial distribution, whereas high-frequency details (such as edges) make up a smaller proportion. During the reconstruction process, neural networks first restore low-frequency information, which is typically easier to recover, before refining high-frequency details. Existing loss functions do not account for this natural progression in neural network optimization. As a result, using static loss functions throughout training can hinder the network's ability to efficiently and effectively restore both smooth regions and fine details.

% Table generated by Excel2LaTeX from sheet 'Sheet2'
\begin{table}[t]
    \centering
    \vspace{0.5em} % Add vertical space before the caption
    \caption{CLIP is capable of perceiving the presence of rain, with the majority of images across all datasets being classified as having rain. Interestingly and reasonably, the Rain100H~\cite{Rain100} dataset consists almost entirely of severe rain, with 99.5\% of images identified by CLIP. Rain100L~\cite{Rain100}, on the other hand, features weaker rain effects, resulting in some images (approximately 7.5\%) being classified by CLIP as rain-free.}
    \vspace{0.5em} % Add vertical space after the caption
    \small
    \resizebox{0.6\textwidth}{!}{
      \begin{tabular}{lcc}
      \toprule
      \noalign{\vskip 7pt}  
      {Dataset$\backslash$Prompt} & \begin{tabular}[c]{@{}c@{}}\footnotesize\texttt{The Image is Unnatural with} \\ \footnotesize\texttt{Rain Effect and of Poor Quality,} \\ \footnotesize\texttt{Adding to the Image Distortion.}\end{tabular} & \begin{tabular}[c]{@{}c@{}}\footnotesize\texttt{Almost No Rain Effect or} \\ \footnotesize\texttt{Raindrops or Any Distortion.}\end{tabular} \bigstrut[b]\\
      \noalign{\vskip 7pt}  
      \hline
      Rain800~\cite{Rain800} & 94.86 & 5.14 \bigstrut[t]\\
      Rain100H~\cite{Rain100} & 99.50 & 0.50 \bigstrut[t]\\
      Rain100L~\cite{Rain100} & 92.50 & 7.50 \bigstrut[t]\\
      Mixed & 97.78 & 2.22 \bigstrut[t]\\
      \bottomrule
      \end{tabular}}%
    \label{tab:intro-clip-1}%
  \end{table}%

  % Table generated by Excel2LaTeX from sheet 'Sheet2'
\begin{table}[t]
    \centering
    \vspace{0.5em} % Add vertical space before the caption
    \caption{CLIP demonstrates the ability to perceive varying degrees of rain intensity across different datasets. Notably, Rain100H~\cite{Rain100} contains a higher proportion of heavy rain (56.11\%), reflecting its design for severe rain conditions. In contrast, Rain100L~\cite{Rain100} exhibits a distinct pattern, with light rain (46.00\%) being the most prominent and moderate rain (12.50\%) also present, consistent with its focus on lighter rain effects.}
    \vspace{0.5em} % Add vertical space after the caption
    \small
    \resizebox{\textwidth}{!}{
    \begin{tabular}{lccc}
    \toprule
    \noalign{\vskip 7pt}  
    {Dataset$\backslash$Prompt} & 
    \begin{tabular}[c]{@{}c@{}}\footnotesize\texttt{A Light Drizzle with Fine Raindrops,} \\ \footnotesize\texttt{Causing Minimal Disruption to Visibility} \\ \footnotesize\texttt{and Surface Accumulation.}\end{tabular} & 
    \begin{tabular}[c]{@{}c@{}}\footnotesize\texttt{A Steady Rainfall with Moderately} \\ \footnotesize\texttt{Sized Raindrops, Leading to Consistent} \\ \footnotesize\texttt{Water Accumulation and Noticeable} \\ \footnotesize\texttt{Reduction in Visibility.}\end{tabular} & 
    \begin{tabular}[c]{@{}c@{}}\footnotesize\texttt{A Heavy Downpour with Large,} \\ \footnotesize\texttt{Dense Raindrops, Resulting in} \\ \footnotesize\texttt{Significant Water Accumulation} \\ \footnotesize\texttt{and Severely Reduced Visibility.}\end{tabular} \bigstrut[b]\\
      \noalign{\vskip 7pt}  
      \hline
      Rain800~\cite{Rain800} & 41.86 & 20.71 & 37.43 \bigstrut[t]\\
      Rain100H~\cite{Rain100} & 37.94 & 5.940 & 56.11 \bigstrut[t]\\
      Rain100L~\cite{Rain100} & 46.00 & 12.50 & 41.50 \bigstrut[t]\\
      Mixed & 39.56 & 10.26 & 50.19 \bigstrut[t]\\
      \bottomrule
      \end{tabular}}%
    \label{tab:intro-clip-2}%
  \end{table}%

To overcome these challenges, we propose a novel approach that combines the power of contrastive language-image pre-training (CLIP)~\cite{CLIP} with a rain perception network (RPN) and mask-guided cross-attention (MGCA) mechanisms. As demonstrated in \cref{tab:intro-clip-1,tab:intro-clip-2}, CLIP shows remarkable capability in detecting the presence and intensity of rain across different datasets. This quantitative analysis validates CLIP's capability to distinguish between different rain intensities, which is crucial for developing adaptive deraining algorithms. Specifically, \cref{tab:intro-clip-1} presents the probability of images across the commonly used datasets being classified into two categories: images with rain effects and those without. The results show that Rain100H~\cite{Rain100} has the highest percentage (99.5\%) of images classified as containing rain, followed by the mixed data of the three datasets (97.78\%), then Rain800~\cite{Rain800} (94.86\%), and Rain100L~\cite{Rain100} (92.5\%). This indicates that Rain100H~\cite{Rain100} contains the most severe rain effects, while Rain100L~\cite{Rain100} includes some images with minimal rain that are classified as rain-free. \cref{tab:intro-clip-2} further breaks down the rain intensity into three levels: light, moderate, and heavy. In the Rain100H~\cite{Rain100} dataset, heavy rain is the most prevalent (56.11\%), aligning with its purpose of representing severe rain scenarios. Conversely, Rain100L~\cite{Rain100} is characterized by a predominance of light and moderate rain (58.50\%), highlighting its emphasis on milder rain conditions. Our approach leverages CLIP's ability to align images with textual descriptions to guide the network in perceiving different rain patterns. This enables the model to dynamically adjust to various types of rain, improving performance in complex scenarios. Additionally, we address the issue of interaction between rainy and non-rainy regions by introducing a mask-guided cross-attention mechanism, which enhances feature extraction and interaction across these regions, ensuring that both rainy and non-rainy areas are effectively processed for better restoration. We also propose a dynamic loss scheduling strategy that emphasizes low-frequency regions in the early stages of training and gradually shifts focus to high-frequency details as training progresses, which improves convergence and enables more effective restoration of both coarse and fine details.

The key contributions of this paper are as follows:
\begin{itemize}
    \item We introduce a CLIP-driven rain perception module that enables the model to detect different rain patterns through semantic text-image alignment. This module allows the model to select the most appropriate processing path based on the detected rain type, improving the model’s adaptability to a wide range of rain conditions.
    \item We propose a mask-guided cross-attention mechanism that enhances feature extraction and interaction between rainy and non-rainy regions. By effectively guiding the restoration process, this mechanism ensures that the network can remove rain from affected areas while preserving the details in clear, non-rainy regions.
    \item We develop a dynamic loss scheduling strategy that adjusts the model’s attention during training. In the early stages, the loss function focuses on low-frequency smooth regions, which are easier to restore, while in the later stages, it prioritizes high-frequency details, which are harder to recover. This dynamic adjustment improves the network’s convergence and overall performance in image restoration tasks.
    \item A comprehensive set of experiments were conducted, including comparison with state-of-the-art methods, ablation studies, parameter analysis, and visualizations. These experiments demonstrate the superiority of our model, providing empirical evidence of its effectiveness in handling complex and diverse rain patterns.
\end{itemize}

The remainder of the paper is organized as follows: In Section \ref{sec:related}, we review the related work on image deraining, particularly focusing on the challenges and advancements in recent methods. Section \ref{sec:method} presents the detailed methodology of our proposed approach. Section \ref{sec:experiments} discusses the experimental setup, including datasets, evaluation metrics, and results. Finally, Section \ref{sec:conclusion} concludes the paper and outlines potential future work.

\section{Related Work}\label{sec:related}

\subsection{Single Image Deraining}

Traditional methods for single image deraining primarily rely on model-driven approaches, including filter-based and priori-based methods~\cite{Li_2022_CVPR,10945649,ZHANG2023109740}. The filter-based methods focus on physical filtering mechanisms to reconstruct clean images~\cite{6272780}, whereas priori-based methods formulate rain removal as an optimization problem by leveraging sparse priors~\cite{5946766}, Gaussian Mixture Models (GMM)~\cite{6126278,Li_2016_CVPR}, and low-rank representations~\cite{du2018single}. Despite their theoretical foundations, these conventional methods exhibit significant limitations in addressing complex rain patterns characterized by diverse sizes, orientations, and densities of rain streaks~\cite{10945649,yang2020single}.

In recent years, with methodologies gradually shifting from model-driven to data-driven ones, deep learning (DL)-based approaches have made substantial progress in the field of single image deraining~\cite{su2023survey,ZHANG2023109740}. Early DL-based approaches primarily utilized convolutional neural networks (CNNs), where foundational works like DerainNet~\cite{DerainNet} established CNN-based frameworks for learning rain-to-clean image mappings. Subsequent advancements incorporated more sophisticated CNN architectures, including residual networks for enhanced feature learning~\cite{DDN,DAF-Net} and U-Net variants for improved hierarchical processing~\cite{HCN,Syn2Real}, \etc To further improve the fidelity of restored images, generative adversarial networks (GANs)~\cite{GANs} were adopted by leveraging adversarial training and unsupervised learning techniques~\cite{NLCL,DCD-GAN}. However, the inherent spatial invariance and localized receptive fields of convolutional operations restrict the ability of CNNs and GANs to model spatially variant rain patterns and global image structures. To address these limitations, Transformers~\cite{Attention} emerged as powerful alternatives by capturing long-range dependencies and non-local patterns critical for better image reconstruction~\cite{DRSformer,chen2024bidirectional,chen2024rethinking}. Inspired by the recent success of diffusion models~\cite{diffusion} in generating high-quality images, diffusion-based approaches such as WeatherDiffusion~\cite{WeatherDiffusion}, RDDM~\cite{RDDM}, and RainDiffusion~\cite{Raindiffusion} have shown promising results in various image restoration tasks, potentially advancing image deraining. Nevertheless, due to the quadratic time complexity of Transformers~\cite{Attention} and the sampling strategies of diffusion models~\cite{diffusion}, researchers have attempted to use linear Mamba~\cite{mamba,mamba2,MS-RainMamba} models to improve the efficiency of image generation while maintaining the quality of generated images.

Despite significant advancements, the performance of existing deraining models may be limited when encountering complex and mixed rain patterns. Therefore, this paper proposes the use of CLIP~\cite{CLIP} to perceive different types of rains, allowing for the specialized handling of various intricate rainy scenarios.

\subsection{CLIP for Image and Text Matching}

Contrastive Language-Image Pre-training (CLIP)~\cite{CLIP} is a multimodal framework that aligns visual and textual representations through contrastive learning. It has been widely used in various tasks, including image classification, retrieval, and generation, \etc CLIP employs dual encoders: a vision encoder (\eg, Vision Transformer~\cite{ViT} or ResNet~\cite{ResNet}) and a text encoder (Transformer-based~\cite{Attention}), trained on 400 million image-text pairs. The model optimizes a contrastive loss to maximize similarity between matched image-text pairs while minimizing similarity for mismatched pairs, effectively projecting both modalities into a shared embedding space. The alignment mechanism and scalability of CLIP have established it as a foundational tool for cross-modal understanding.

CLIP's versatility has spurred innovations across various computer vision tasks. For zero-shot image classification, CLIP achieves performance comparable to supervised models, where CLIP can classify images by comparing their embeddings with text prompts of target categories, bypassing task-specific fine-tuning. In cross-modal retrieval, CLIP-Forge~\cite{sanghi2022clip} utilizes CLIP's joint embeddings to enable text-to-image and image-to-text searches with high accuracy. CLIP has also been adapted for dense prediction tasks. For instance, DenseCLIP~\cite{rao2022denseclip} integrates CLIP with segmentation frameworks by aligning regional visual features with textual descriptions. Additionally, the generalizability of CLIP has been exploited for few-shot learning, where methods like Tip-Adapter~\cite{zhang2021tip} refine CLIP's embeddings with minimal labeled data. These applications highlight CLIP's ability to bridge semantic gaps between vision and language.

CLIP has also been explored for low-level vision tasks. For instance, DiffusionCLIP~\cite{kim2022diffusionclip} integrates CLIP with diffusion models to enable text-guided image manipulation. LDFusion~\cite{wang2024infrared} leverages CLIP's perceptual capabilities to learn semantic shifts between infrared and RGB images, thereby adaptively fusing complementary information from both modalities. DA-CLIP~\cite{luo2023controlling} transfers CLIP to low-level vision tasks as a multi-task framework for universal image restoration. Drawing inspiration from these CLIP models, this paper explores the application of CLIP for rain pattern perception to enhance adaptability and robustness in handling diverse conditions. We propose leveraging CLIP's multimodal alignment capabilities to recognize and characterize various rainfall patterns (\eg, heavy downpour, light drizzle, directional rain) through semantic text-image correlation. Our approach enables dynamic adaptation to different rain types without requiring explicit physical modeling, potentially advancing deraining performance across diverse rainy conditions.

\section{Method}\label{sec:method}

\begin{figure}[H]
    \centering
    \includegraphics[width=1\linewidth]{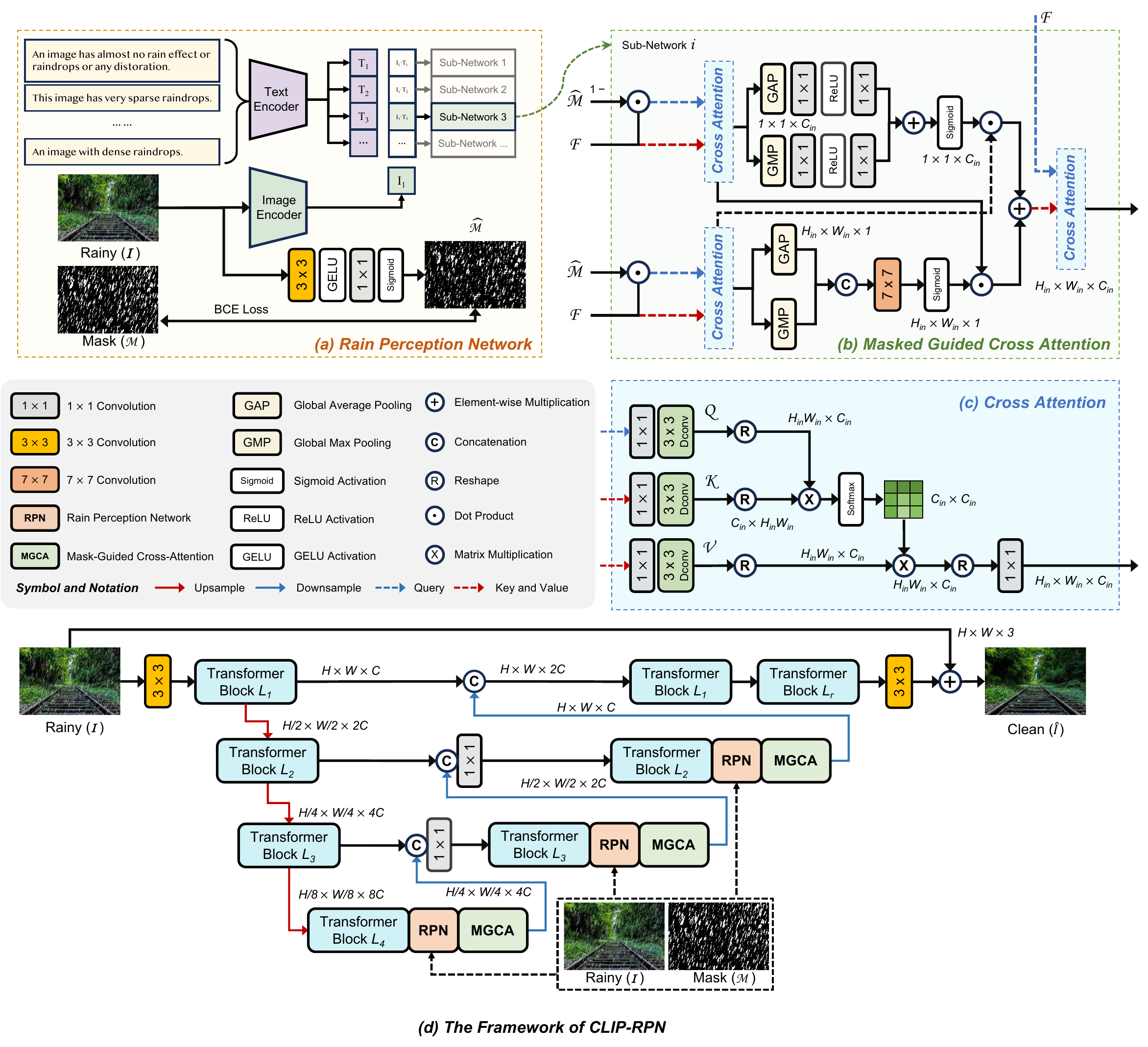}
    \caption{Overview of the proposed method}
    \label{fig:overview}
\end{figure}

\subsection{Overview}

The overall structure of the proposed network (CLIP-RPN), along with its detailed components, is illustrated in Fig~\ref{fig:overview}. As shown in Fig.~\ref{fig:overview}(d), the network follows a U-shaped~\cite{UNet} architecture with downsampling and skip connections. CLIP-RPN has four feature extraction layers, where the spatial resolution decreases and the number of channels increases as the depth increases. During reconstruction, features from intermediate layers are upsampled and concatenated with the original features, and the final clean image is obtained by adding the learned residuals to the rainy input image. Vision Transformer (ViT)~\cite{ViT} is employed in each layer for feature extraction via self-attention. Before each upsampling step, we integrate the proposed Rain Perception Network (RPN) and Mask-Guided Cross-Attention (MGCA) network to enhance the network's ability to perceive rain patterns and interact with features  from different regions of the image.

\subsection{CLIP-Driven Rain Perception}

As shown in Fig.~\ref{fig:overview}(a), the CLIP-Driven Rain Perception (RPN) module is responsible for guiding the network to route to the appropriate sub-network based on different rain patterns. To achieve this, we design various text prompts that help the network select the most suitable sub-network for processing. This enables the network to choose the right processing branch based on the detected rain patterns, enhancing its robustness and reducing the load on a single network.

In the RPN , we use the pre-trained CLIP model to extract features from the input rainy image and text prompts. Specifically, we input the image $\mathbf{I} \in \mathbb{R}^{H \times W \times 3}$ (where $H$ and $W$ are the height and width of the image) into CLIP's image encoder $\Phi_{v}$ (using the ViT-B/32 architecture), obtaining the image feature vector $\mathbf{I}_1 \in \mathbb{R}^{1 \times 512}$. Similarly, we input a set of text prompts $\mathbf{P} = \{\mathbf{P}_1, \mathbf{P}_2, \cdots, \mathbf{P}_n\}, i = 1, 2, \cdots, n$ into CLIP's text encoder $\Phi_{t}$ to obtain their corresponding textual feature vectors $\mathbf{T}_i \in \mathbb{R}^{1 \times 512}, i = 1, 2, \cdots, n$. This process is represented as:
\begin{equation}
    \mathbf{I}_1 = {\Phi_{v}}(\mathbf{I}), \quad \mathbf{T}_i = {\Phi_{t}}(\mathbf{P}_i), \quad i \in {1, 2, \cdots, n}
\end{equation}

To leverage CLIP's perceptual capabilities, we calculate the similarity between the image feature $\mathbf{I}_1$ and each text feature $\mathbf{T}_i$, obtaining the similarity scores $s_i$. The index $s$ corresponding to the maximum similarity score is then used to route the image $\mathbf{I}$ to the sub-network with index $s$, as follows:
\begin{equation}
\begin{aligned}
    s_i &=  \frac{e^{\mathbf{I}_1 \cdot \mathbf{T}_i}}{\sum_{j=1}^{n} e^{\mathbf{I}_1 \cdot \mathbf{T}_j}}, \quad i = 1, 2, \dots, n \\
    s &= \text{argmax}(s_i), \quad i = {1, 2, \cdots, n}
\end{aligned}
\end{equation}

The CLIP-driven rain perception method provides global guidance, directing the network to the most suitable sub-network for processing based on the detected rain patterns. However, we also require fine-grained pixel-wise perception. Therefore, in addition to CLIP's global semantic perception, we incorporate pixel-level rain drop prediction. Specifically, a simple convolutional neural network is used to predict the raindrop density at each pixel, which serves as a weight for guiding feature extraction. This process is expressed as:
\begin{equation}
    \hat{\mathbf{M}} = \sigma(W_2^{1\times 1}(\delta(W_1^{3\times 3}(\mathbf{I}))))
\end{equation}
where $\hat{\mathbf{M}}$ is the predicted raindrop confidence, $W_1^{3\times 3}$ and $W_2^{1\times 1}$ are convolutional layers with kernel size of $3\times 3$ and $1\times 1$, respectively. $\sigma$ and $\delta$ are activation functions (Sigmoid and GELU, respectively). We compute the absolute difference between the original image and the ground truth image, treating pixels with an absolute difference greater than $0.1$ as containing raindrops, resulting in a binary mask $\mathbf{M}$. The binary rain mask $\mathbf{M}$ is generated as:
\begin{equation}
    \mathbf{M} = \begin{cases}
        1, & \text{if } |\mathbf{I} - \mathbf{I}'| > 0.1 \\
        0, & \text{otherwise}
    \end{cases}
\end{equation}

The predicted rain mask $\hat{\mathbf{M}}$ is supervised using a binary cross-entropy loss against ground truth $\mathbf{M}$:
\begin{equation}
    l_{bce} = -\frac{1}{H \times W} \sum_{i=1}^{H} \sum_{j=1}^{W} \left( \mathbf{M}_{i,j} \log(\hat{\mathbf{M}}_{i,j}) + (1 - \mathbf{M}_{i,j}) \log(1 - \hat{\mathbf{M}}_{i,j}) \right)
\end{equation}

After obtaining the global and fine-grained rain patterns and confidence map, we route the image $\mathbf{I}$ to the sub-network indexed by $s$ for processing. This sub-network receives both the raindrop confidence $\hat{\mathbf{M}}$ and the features $\mathbf{F}$ extracted from the previous Transformer block as inputs, yielding the features $\mathbf{F}_s$ after interaction between rainy and non-rainy regions:
\begin{equation}
    \mathbf{F}_s = \Psi_s(\mathbf{F}, \hat{\mathbf{M}})
\end{equation}
where $\Psi_s$ is the $s$-th sub-network with the mask-guided cross-attention, which will be elaborated in details in the next section.

\subsection{Mask-Guided Cross-Attention}

As shown in Fig.~\ref{fig:overview}(b), the Mask-Guided Cross-Attention (MGCA) network is designed to address the differences in feature representations between rainy and non-rainy regions in the image. The features of rainy regions tend to contain more noise and less details, while the non-rainy regions generally retain clearer image details. Therefore, extracting the features for each region separately helps to effectively perform feature selection and weighting, optimizing the network's perception of rainy and non-rainy areas. Given the features $\mathbf{F}$ extracted from the last layer of the Transformer block and the raindrop confidence map $\hat{\mathbf{M}}$, for the $s$-th MGCA, we first obtain the features of both rainy and non-rainy regions:
\begin{equation}
    \begin{aligned}
        \mathbf{F}_r &= \mathbf{F} \odot \hat{\mathbf{M}} \\
        \mathbf{F}_n &= \mathbf{F} \odot (1 - \hat{\mathbf{M}})
    \end{aligned}
\end{equation}
where $\odot$ represents element-wise multiplication, with $\mathbf{F}_r$ and $\mathbf{F}_n$ denoting the features of the rainy and non-rainy regions, respectively. Next, we input $\mathbf{F}_r$ and $\mathbf{F}_n$ into two independent multi-dconv head transposed cross-attention modules as queries, while using the original features $\mathbf{F}$ as keys and values. This step is aimed at identifying and extracting different feature parts, guided by the rain or non-rain location. The structure of the cross-attention module is shown in Fig~\ref{fig:overview}(c). For a given query feature $\mathbf{F_q}$, key feature $\mathbf{F_k}$, and value feature $\mathbf{F_v}$, the cross-attention operation (CA) is defined as follows.
\begin{equation}
    \begin{aligned}
        \mathbf{Q} &= DW_q^{3\times 3}(W_q^{1\times 1}(\mathbf{F_q})) \\
        \mathbf{K} &= DW_k^{3\times 3}(W_k^{1\times 1}(\mathbf{F_k})) \\
        \mathbf{V} &= DW_v^{3\times 3}(W_v^{1\times 1}(\mathbf{F_v}))
    \end{aligned}
\end{equation}

First, $\mathbf{Q}$, $\mathbf{K}$, and $\mathbf{V}$ are obtained by applying a $1 \times 1$ convolution followed by a $3 \times 3$ depth-wise convolution to project the features. The depth-wise convolution layers, $DW_q^{3\times 3}$, $DW_k^{3\times 3}$, and $DW_v^{3\times 3}$, use kernel sizes of $3 \times 3$. Then, we compute the similarity between $\mathbf{Q}$ and $\mathbf{K}$ to obtain the attention weights and multiply them by $\mathbf{V}$ to produce the final feature map $\mathbf{F}_{attn}$:
\begin{equation}
    \mathbf{F}_{attn} = \text{softmax}(\frac{\mathbf{Q} \mathbf{K}^T}{\alpha})\mathbf{V}
\end{equation}
where $\alpha$ is a trainable scaling parameter employed to modulate the dot production.

We then apply CA to both the rainy and non-rainy regions, resulting in query features for each:
\begin{equation}
    \mathbf{F}_n^{ca} = \text{CA}(\mathbf{F_n}, \mathbf{F}, \mathbf{F}),
    \mathbf{F}_r^{ca} = \text{CA}(\mathbf{F_r}, \mathbf{F}, \mathbf{F})
\end{equation}

To further promote feature interaction between rainy and non-rainy regions, we extract the importance of the non-rainy region's channels and multiply them with features of the rainy areas. For the rainy regions, we extract the spatial pixel importance and multiply it with the non-rainy areas.

\textbf{Non-rainy} regions generally contain more distinct image information, making it easier to learn the importance of each channel based on the known features. Conversely, non-rainy regions may lack accurate modeling of raindrop regions. By extracting channel importance from the non-rainy region and interacting it with the rainy features, the network can effectively guide the processing of rainy features using the clear features from the non-rainy regions, thereby helping to remove noise and restore the impact of rain, which is formulated as:
\begin{equation}
    \begin{aligned}
    \mathbf{F}_n^{ca} &\leftarrow \mathbf{F}_n^{ca} \odot \mathbf{W}_r, \quad\quad \text{where} \\
    \mathbf{W}_r &= \sigma(\text{Cat}(\mathbf{F}_{r}^{avg}, \mathbf{F}_{r}^{max}))
    \end{aligned}
\end{equation}
where $\mathbf{F}_{r}^{avg}$ and $\mathbf{F}_{r}^{max}$ represent the average pooling and max pooling results of $\mathbf{F}_r^{ca}$, respectively. The $\text{Cat}(\cdot)$ operation denotes concatenation.

\textbf{Rainy} regions may contain noise or occlusions that could affect visual quality, and these noisy features are typically localized in space. By extracting the spatial pixel importance from the rainy region and interacting it with the non-rainy features, the network can learn how to preserve the clear information of the non-rainy regions while removing the rain in the rainy regions, minimizing unnecessary effects from the rain areas during the image restoration process. This operation is defined as:
\begin{equation}
    \begin{aligned}
    \mathbf{F}_r^{ca} &\leftarrow \mathbf{F}_r^{ca} \odot \mathbf{W}_n, \quad\quad \text{where} \\
    \mathbf{W}_n &= \sigma\left( W^{1 \times 1}_2(\gamma(W^{1 \times 1}_1(\mathbf{F}_{n}^{avg})))  +  W^{1 \times 1}_4(\gamma(W^{1 \times 1}_3(\mathbf{F}_{n}^{max}))) \right)
    \end{aligned}
\end{equation}
where $\mathbf{F}_{n}^{avg}$ and $\mathbf{F}_{n}^{max}$ represent the average and max pooling results of $\mathbf{F}_n^{ca}$, respectively. $\gamma$ is the ReLU activation function, and $W^{1 \times 1}_i$, $i = {1, 2, 3, 4}$, represent the convolutional layers with kernel size $1 \times 1$. Finally, the interaction-completed rainy region features $\mathbf{F}_r^{ca}$ and non-rainy region features $\mathbf{F}_n^{ca}$ are summed to produce the final features $\mathbf{F}_s$:
\begin{equation}
    \mathbf{F}_s = \mathbf{F}_r^{ca} + \mathbf{F}_n^{ca}
\end{equation}

To further refine the fused features, we use the original features to generate queries, while the interaction-completed features $\mathbf{F}_s$ are used to generate keys and values. A final cross-attention operation is performed to obtain the output features $\mathbf{F}_{out}$. The mask-guided cross-attention achieves a fine-grained interaction of features from rainy and non-rainy regions through multiple cross-attention layers guided by the predicted rain confidence map, enhancing the network's feature extraction capabilities. Each MGCA sub-network will process the raw features routed from the previous step that correspond to specific rain patterns, thereby improving the network's compatibility with different rain modes through this divide-and-conquer strategy.

\subsection{Dynamic Loss Scheduling}
\label{sec:dynamic_loss_scheduling}

\begin{figure}[H]
    \centering
    \includegraphics[width=0.8\linewidth]{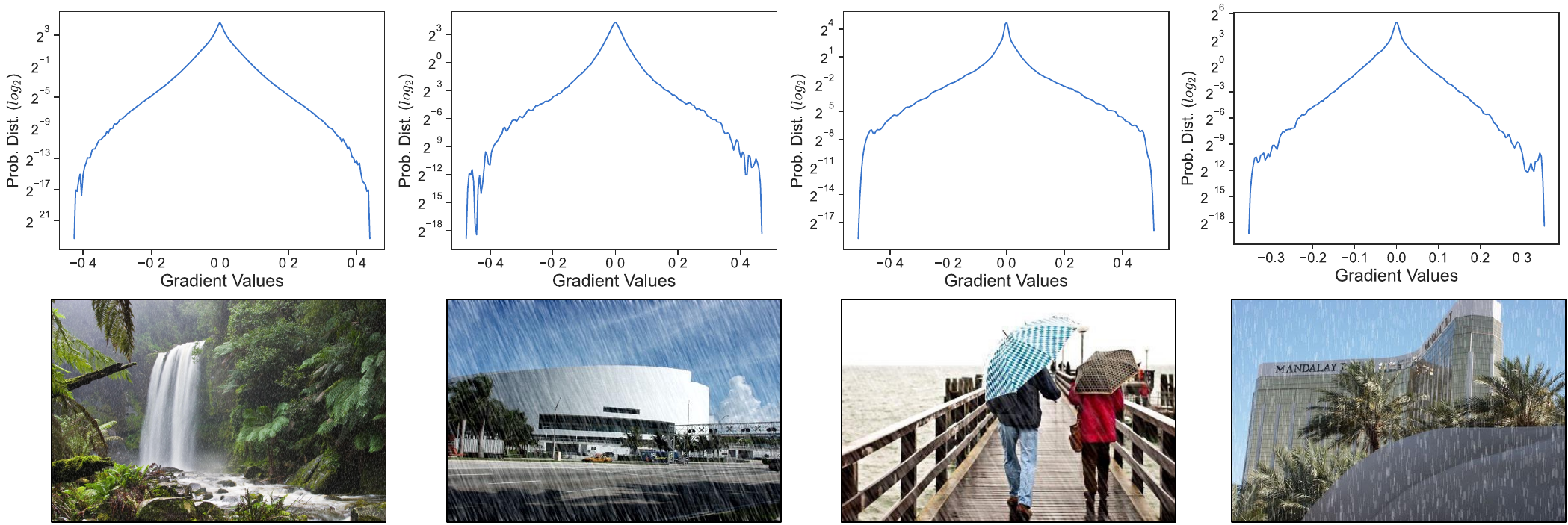}
    \caption{The logarithmic gradients distribution (the second row) of natural images (the first row).}\label{fig:grad}
\end{figure}

\begin{figure}[t]
    \hspace{-5mm}
     \centering
     \subfloat
     {\includegraphics[width=0.2\textwidth]{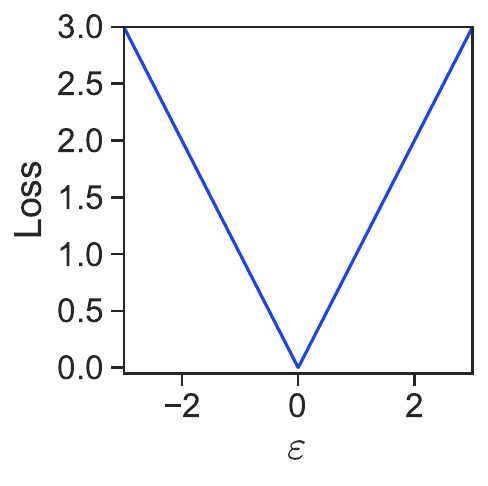}}
     \subfloat
     {\includegraphics[width=0.2\textwidth]{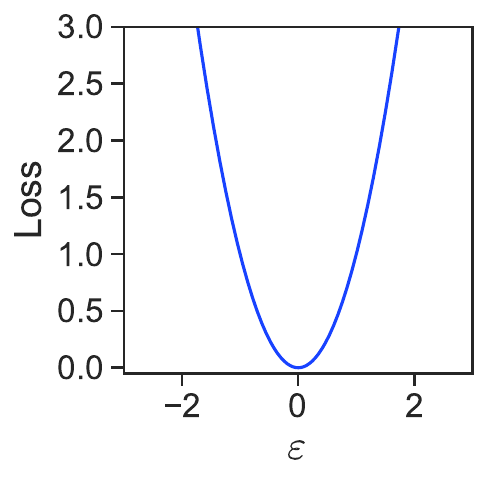}}
     \subfloat
     {\includegraphics[width=0.2\textwidth]{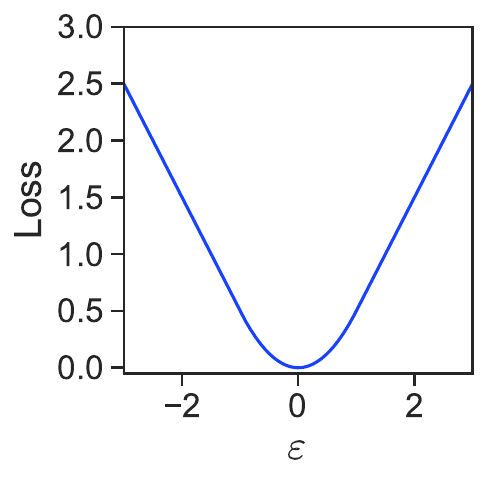}}
  
     \addtocounter{subfigure}{-3} % Reset subfigure counter for the next set of subfloats
     \vspace{-3.5mm}
     \hspace{-5mm}
     \subfloat[$\ell_1$ loss]
     {\includegraphics[width=0.2\textwidth]{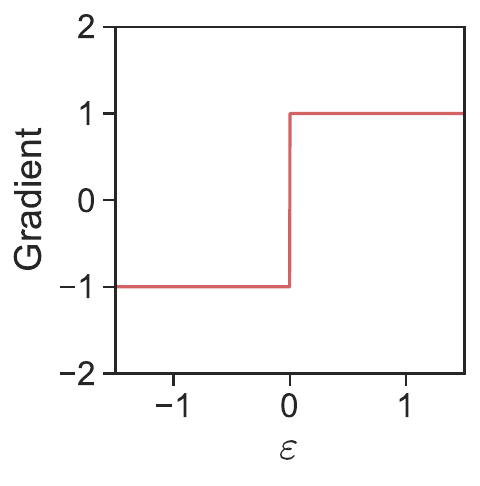}\label{fig:l1grad}}
     \subfloat[$\ell_2$ loss]
     {\includegraphics[width=0.2\textwidth]{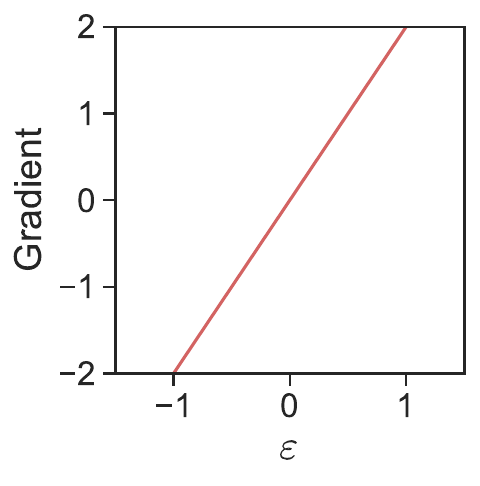}\label{fig:l2grad}}
     \subfloat[Huber loss]
     {\includegraphics[width=0.2\textwidth]{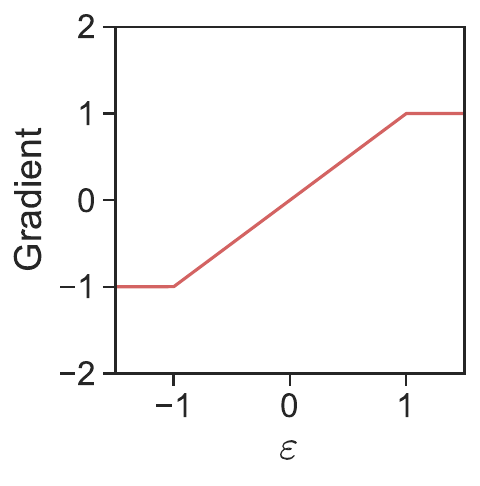}\label{fig:hubergrad}}
  
     \caption{The commonly used loss functions and their gradients with respect to the prediction error.}\label{fig:lossgrad}
   \end{figure}

   \begin{figure}[t]    
     \centering
     \subfloat[]
     {\hspace{2mm}\includegraphics[width=0.3\textwidth]{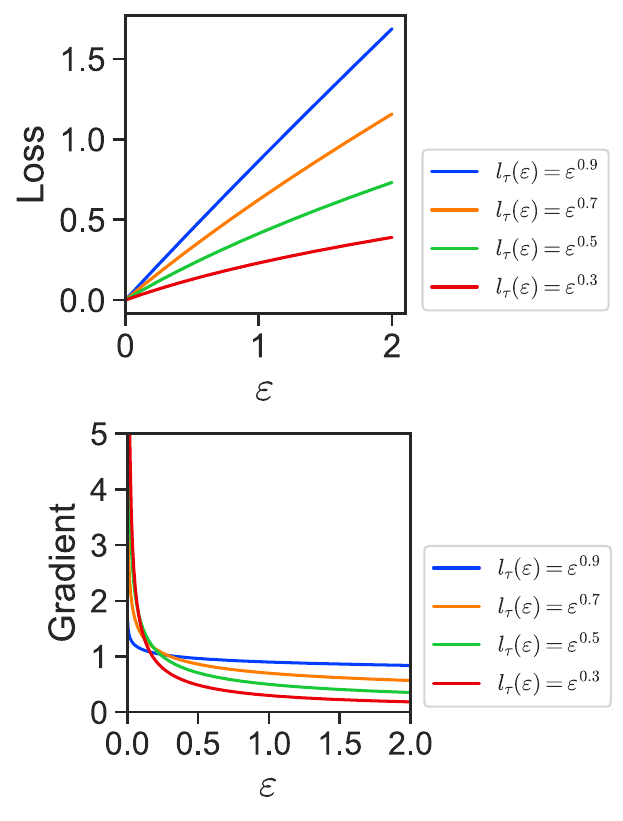}\label{fig:dpl1}}
     \subfloat[]
     {\hspace{2mm}\includegraphics[width=0.3\textwidth]{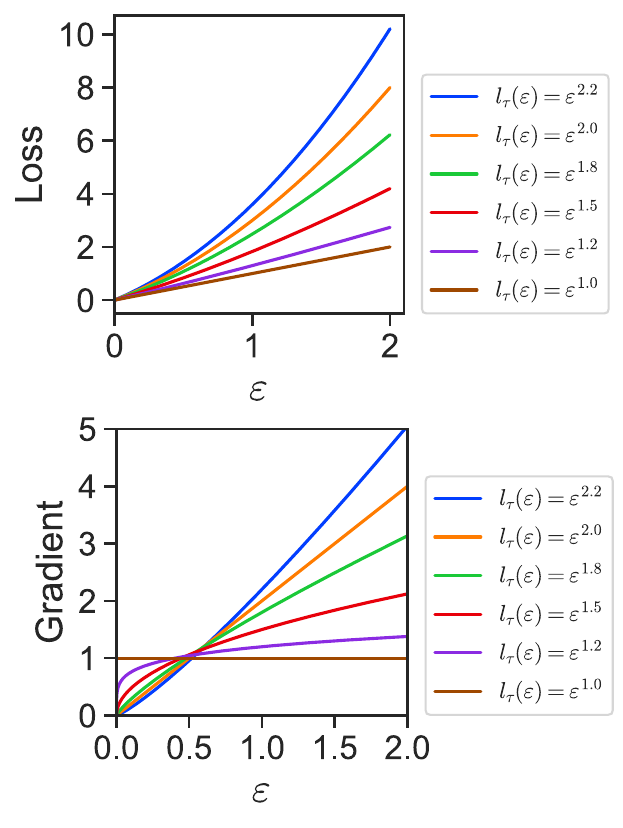}\label{fig:dpl2}}
     \caption{The proposed loss function and its gradient with respect to the prediction error and the dynamically scheduled exponential parameter.}\label{fig:lossdsl}
   \end{figure}

In the realm of computer vision, regardless of the network structure employed, reconstruction loss stands as a vital component in the optimization of networks, which is used to force the predicted pixels to close to the actual values. Commonly used loss functions are $l_1$ loss, $l_2$ loss, and their combination such as huber loss. Although these loss functions have been proven to be highly effective in the field of computer vision for low-level tasks, we argue that it may not be the best choice.

In natural image statistics, low-frequency smooth areas dominate the spatial distribution while high-frequency details constitute a relatively minor proportion. Our empirical analysis of multiple randomly selected images (Fig.~\ref{fig:grad}) reveals that logarithmic gradient distributions of natural images exhibit a pronounced concentration in low-frequency regions, with high-frequency domains demonstrating sparsity. In addition, prior research in neural network dynamics~\cite{DIP,pmlr-v80-lehtinen18a} has identified a hierarchical learning pattern: networks initially prioritize smoother regions in reconstruction process due to its inherent recoverability, subsequently progressing to high-frequency refinement. During early optimization phases, smoother regions with lower spatial frequencies typically exhibit smaller reconstruction discrepancies, whereas high-frequency regions demonstrate proportionally larger errors.

Intuitively, based on the above analysis, to align with and accelerate this trend in the network for faster convergence, the loss function should focus more on the low-frequency smooth information (\ie, smaller errors) in the early stages and give more attention to high-frequency information (\ie, larger errors) in the later stages. However, as shown in Fig.~\ref{fig:lossgrad}, the $l_1$ loss maintains constant gradient magnitude ($\nabla l_1 = 1 $) independent of error scale, enforcing equal attention across all pixels throughout training. Conversely, $l_2$ loss produces error-proportional gradients ($\nabla l_2 \propto \varepsilon$), creating an inverse prioritization where larger initial errors in high-frequency regions disproportionately dominate early optimization stages. Clearly, neither the $l_1$ loss nor the $l_2$ loss adequately reflects the gradient distribution characteristics of natural images or aligns well with the network optimization process.

To address this, we propose a dynamic loss scheduling strategy, which allocates more attention to smoother region in the early stages of the network and more focus on larger error area in the later stages. Specifically, for an error $\varepsilon$, the loss is produced as:
\begin{equation}
    l_{\tau}(\varepsilon) = \varepsilon^{\beta+\eta f(\tau)/\text{T}},
\label{eq:dynamic_loss_scheduling}
\end{equation}
where $\tau={0, 1, \dots, \text{T}}$. $\beta$ and $\eta$ are hyperparameters that jointly determine the gradient range from start to end. $\text{T}$ is a constant determined after specifying the training dataset size, epoch, and batch size.  $f(\tau)$ is the scheduling function, which we set as linear, \ie, $f(\tau) = \tau$.

Fig.~\ref{fig:lossdsl} graphically presents the loss function defined in Eq.~(\ref{eq:dynamic_loss_scheduling}) along with its gradient characteristics. As shown, when the exponential parameter of Eq.~(\ref{eq:dynamic_loss_scheduling}) is less than $1$ (Fig.~\subref*{fig:dpl1}), the loss assigns larger gradients to sample pixels with smaller loss values (typically corresponding to low-frequency smooth regions), whereas for exponential greater than $1$ (Fig.~\subref*{fig:dpl2}), the inverse relationship holds. In other words, the parts with an exponential less than $1$ correspond to the early stages of restoring the low-frequency smoother information, while those greater than $1$ correspond to the later stages of reconstructing the high-frequency information. This parametric behavior effectively mimics the natural progression of neural network optimization, thus achieving better results than either the $l_1$ or $l_2$ loss functions.

Combining the binary cross-entropy loss from mask-guided cross-attention, the total loss is set as:
\begin{equation}
    l_{total} = \sum_{i=1}^{3} l_{bce}^{(i)}*0.1 + l_{\tau}
\end{equation}
where $l_{bce}^{(i)}$ is the binary cross-entropy loss from the $i$-th level MGCA and 0.1 is an empirical weight factor.

\section{Experiments}\label{sec:experiments}

This section presents a comprehensive evaluation of our proposed CLIP-RPN method. Through extensive comparative experiments with state-of-the-art methods, we demonstrate the superior performance of our approach across various datasets, particularly in handling complex and diverse rain patterns. Furthermore, we conduct detailed ablation studies and parameter analysis to analyze the contributions of key components.

\subsection{Datasets and Evaluation Metrics}

In our experiments, we evaluate the proposed method on three widely used deraining datasets: (1) Rain100L~\cite{Rain100}, which contains 200 training images and 100 test images with light rain effects; (2) Rain100H~\cite{Rain100}, consisting of 1800 training images and 100 test images with heavy rain conditions; and (3) Rain800~\cite{Rain800}, which includes 700 training images and 100 test images with diverse rain patterns. These datasets provide a comprehensive evaluation of our method's ability to handle different rain intensities and patterns. To further validate our model's capability in handling complex rain patterns, we also conduct experiments on a mixed dataset combining all three datasets, which presents more challenging and diverse rain scenarios. To quantitatively assess the performance of our method, we employ two widely used metrics: Peak Signal-to-Noise Ratio (PSNR) and Structural Similarity Index (SSIM). PSNR measures the pixel-level similarity between the derained image and the ground truth, with higher values indicating better quality. SSIM evaluates the structural similarity between images by considering luminance, contrast, and structure, providing a more perceptually relevant measure of image quality. Both metrics are calculated on the Y channel of the YCbCr color space, following the common practice in image restoration tasks.

\subsection{Implementation Details}

All models were trained on a single NVIDIA RTX 4090 GPU with 24GB memory, with an Intel(R) Xeon(R) Gold 6148 CPU @ 2.40GHz as the processor. We set the batch size to 8 and trained for 300 epochs using mixed precision training to accelerate computation and reduce memory usage. The initial learning rate was set to 0.0002 with the AdamW~\cite{AdamW} optimizer, which had a weight decay of 0.01. The hyperparameters in AdamW~\cite{AdamW} that control the exponential decay rates of the moving averages of the gradient and its square were set to 0.9 and 0.999, respectively. We used a linear warmup scheduler for the first 15 epochs followed by a cosine annealing scheduler for the remaining training. For data augmentation, we applied random cropping to 128 $\times$ 128 patches and used horizontal and vertical flipping with a probability of 0.5. In the comparison experiments, the $\beta$ and $\eta$ in the loss function were set to 0.8 and 2.3, respectively. The default prompt used in the experiments is Prompt 3 from Table~\ref{tab:prompts}.

\subsection{Comparative Experiments}

In this section, we compare our proposed CLIP-RPN model with several state-of-the-art all-in-one image restoration models, including PReNet~\cite{PReNet}, MPRNet~\cite{MPRNet}, AirNet~\cite{AirNet}, Restormer~\cite{Restormer}, TransWeather~\cite{TransWeather}, DRSformer~\cite{DRSformer}, PromptIR~\cite{PromptIR}, and AMIR~\cite{AMIR}. These models are capable of handling complex image restoration scenarios, making them highly comparable to our proposed method.

\subsubsection{Quantitative Analysis}

\begin{table}[htbp]
  \centering
  \caption{Quantitative results of compared models on public datasets and mixed dataset.}
  \small
  \resizebox{0.9\linewidth}{!}{
    \begin{tabular}{lcccccccccc}
      \hline
      \multirow{2}[4]{*}{Model}                                                                   & \multicolumn{2}{c}{Rain100L~\cite{Rain100}} & \multicolumn{2}{c}{Rain100H~\cite{Rain100}} & \multicolumn{2}{c}{Rain800~\cite{Rain800}} & \multicolumn{2}{c}{Mixed} & \multicolumn{2}{c}{Average} \bigstrut                                                                                                                  \\
      \cmidrule(r){2-3} \cmidrule(r){4-5} \cmidrule(r){6-7} \cmidrule(r){8-9} \cmidrule(r){10-11} & PSNR                                        & SSIM                                        & PSNR                                       & SSIM                      & PSNR                                  & SSIM               & PSNR              & SSIM               & PSNR              & SSIM \bigstrut               \\
      \hline
      PReNet~\cite{PReNet}                                                                        & 32.06                                       & 0.8966                                      & 26.32                                      & 0.7733                    & 24.12                                 & 0.7586             & 27.47             & 0.8056             & 27.49             & 0.8085 \bigstrut[t]          \\
      MPRNet~\cite{MPRNet}                                                                        & 32.24                                       & 0.8982                                      & 28.51                                      & 0.8192                    & 26.67                                 & 0.7870             & 31.03             & 0.8515             & 29.61             & 0.8390                       \\
      AirNet~\cite{AirNet}                                                                        & 36.00                                       & 0.9219                                      & 29.44                                      & 0.8374                    & 27.69                                 & 0.8077             & 32.09             & 0.8642             & 31.31             & 0.8578                       \\
      Restormer~\cite{Restormer}                                                                  & 36.10                                       & \underline{0.9221}                          & \underline{29.62}                          & \underline{0.8413}        & \underline{28.22}                     & \textbf{0.8131}    & 32.24             & 0.8648             & \underline{31.55} & \underline{0.8603}           \\
      TransWeather~\cite{TransWeather}                                                            & 27.72                                       & 0.8651                                      & 25.02                                      & 0.7965                    & 24.72                                 & 0.8109             & 27.26             & 0.8573             & 26.18             & 0.8325                       \\
      DRSformer~\cite{DRSformer}                                                                  & 36.15                                       & 0.9191                                      & 29.53                                      & 0.8372                    & 27.88                                 & 0.8117             & 32.22             & 0.8647             & 31.45             & 0.8582                       \\
      PromptIR~\cite{PromptIR}                                                                    & \underline{36.20}                           & \textbf{0.9229}                             & 29.57                                      & 0.8406                    & 27.83                                 & 0.8079             & \underline{32.28} & \underline{0.8655} & 31.47             & 0.8592                       \\
      AMIR~\cite{AMIR}                                                                            & 35.74                                       & 0.9195                                      & 29.50                                      & 0.8384                    & 27.79                                 & 0.8103             & 32.21             & 0.8640             & 31.31             & 0.8580                       \\
      CLIP-RPN (ours)                                                                             & \textbf{36.21}                              & 0.9211                                      & \textbf{29.66}                             & \textbf{0.8423}           & \textbf{28.40}                        & \underline{0.8122} & \textbf{32.42}    & \textbf{0.8671}    & \textbf{31.67}    & \textbf{0.8607} \bigstrut[b] \\
      \hline
    \end{tabular}}%
  \label{tab:cmp-quantitative}%
\end{table}%

The quantitative results are presented in Table~\ref{tab:cmp-quantitative}, where our model achieves competitive performance across all datasets. We observe that the more diverse the rain patterns in the dataset, the greater the improvement of our model compared to other models. This is particularly evident in the Rain800 and Mixed datasets, where our model achieves the most significant performance gains, demonstrating its superior capability in handling complex and varied rain conditions. For instance, in terms of PSNR metric, compared to the second-best model, our approach demonstrates consistent improvements of 0.01, 0.04, 0.18, and 0.14 on the Rain100L, Rain100H, Rain800, and Mixed datasets, respectively. While our model's PSNR scores are generally superior to other models, it is worth noting that in the Rain100L dataset, our SSIM score is slightly lower than Restormer~\cite{Restormer} and PromptIR~\cite{PromptIR}. However, this difference is minimal and does not significantly impact overall performance. More importantly, our model demonstrates remarkable performance on the mixed dataset, achieving the highest PSNR and SSIM score. This indicates that our model has superior capability in handling complex rain conditions that combine various rain patterns, which is a more challenging and realistic scenario. The strong performance on the mixed dataset highlights the robustness and adaptability of our approach in dealing with diverse and complex rain conditions.

\begin{table}[htbp]
  \centering
  \caption{Comparison of parameter count, computational complexity, and inference time across different models.}
  \small
  \resizebox{0.45\linewidth}{!}{
    \begin{tabular}{lccc}
      \hline
      Model                            & Param (M) & MACs (G) & Time (Sec) \bigstrut \\
      \hline
      PReNet~\cite{PReNet}             & 0.17      & 66.25    & 13.81 \bigstrut[t]   \\
      MPRNet~\cite{MPRNet}             & 3.64      & 548.65   & 17.29                \\
      AirNet~\cite{AirNet}             & 5.77      & 301.27   & 55.21                \\
      Restormer~\cite{Restormer}       & 26.10     & 140.99   & 26.07                \\
      TransWeather~\cite{TransWeather} & 37.68     & 6.13     & 8.61                 \\
      DRSformer~\cite{DRSformer}       & 33.63     & 220.38   & 69.57                \\
      PromptIR~\cite{PromptIR}         & 32.97     & 158.14   & 27.53                \\
      AMIR~\cite{AMIR}                 & 23.50     & 127.21   & 37.86                \\
      CLIP-RPN (ours)                  & 32.72     & 293.53   & 76.90 \bigstrut[b]   \\
      \hline
    \end{tabular}%
    \label{tab:cmp-size}}%
\end{table}%

\begin{figure}[htbp]
  \centering
  \subfloat
  {\includegraphics[width=\linewidth]{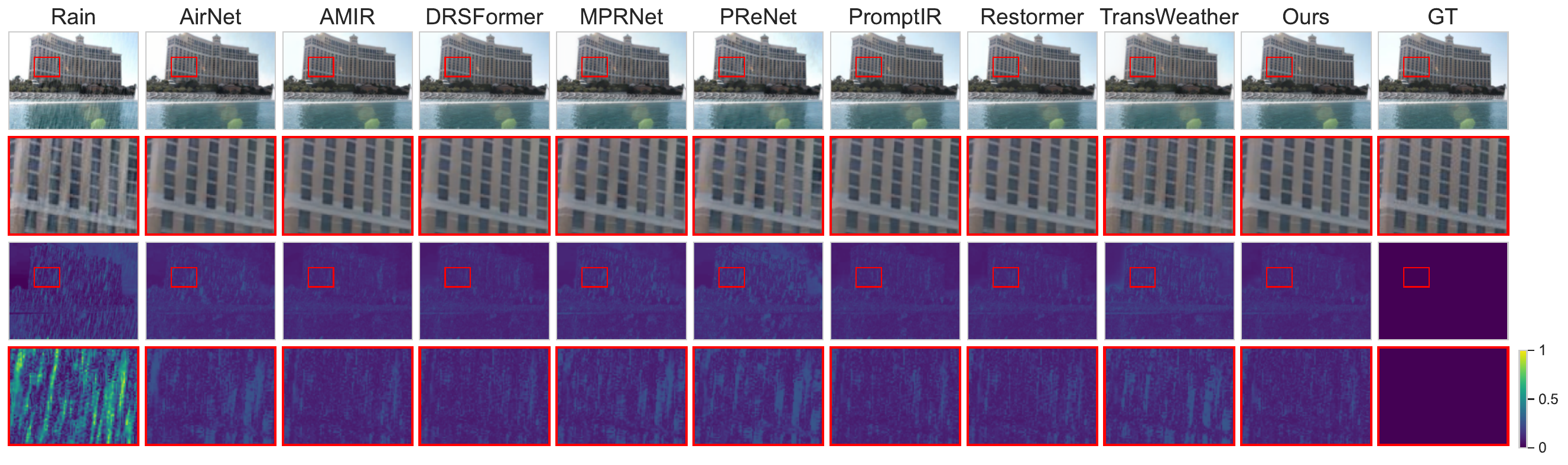}}
  \vspace{-0.5cm}

  \subfloat
  {\includegraphics[width=\linewidth]{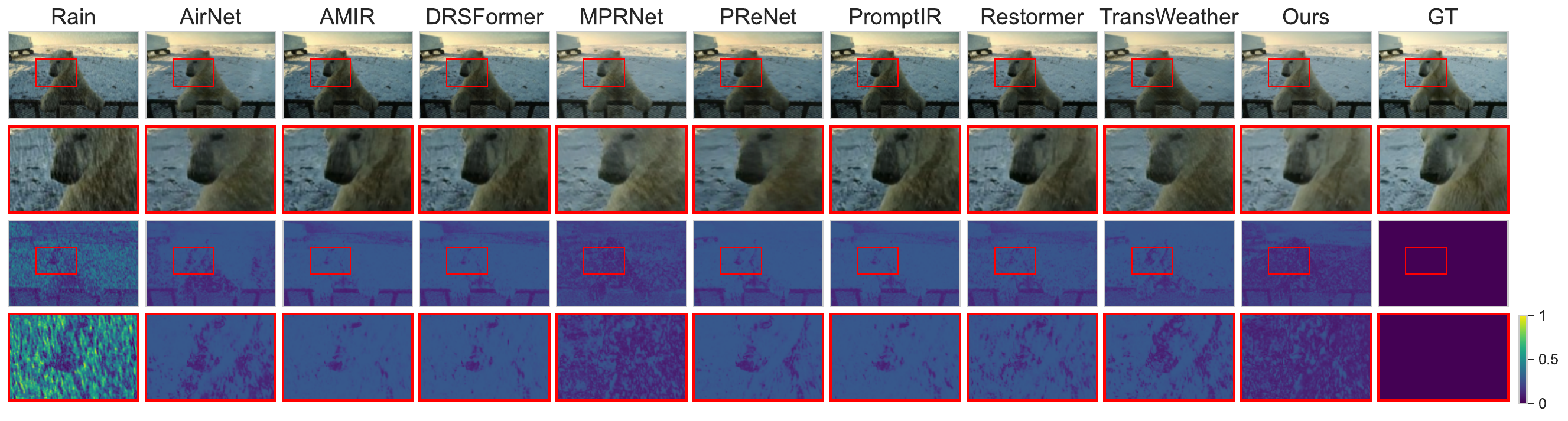}}
  \vspace{-0.5cm}

  \subfloat
  {\includegraphics[width=\linewidth]{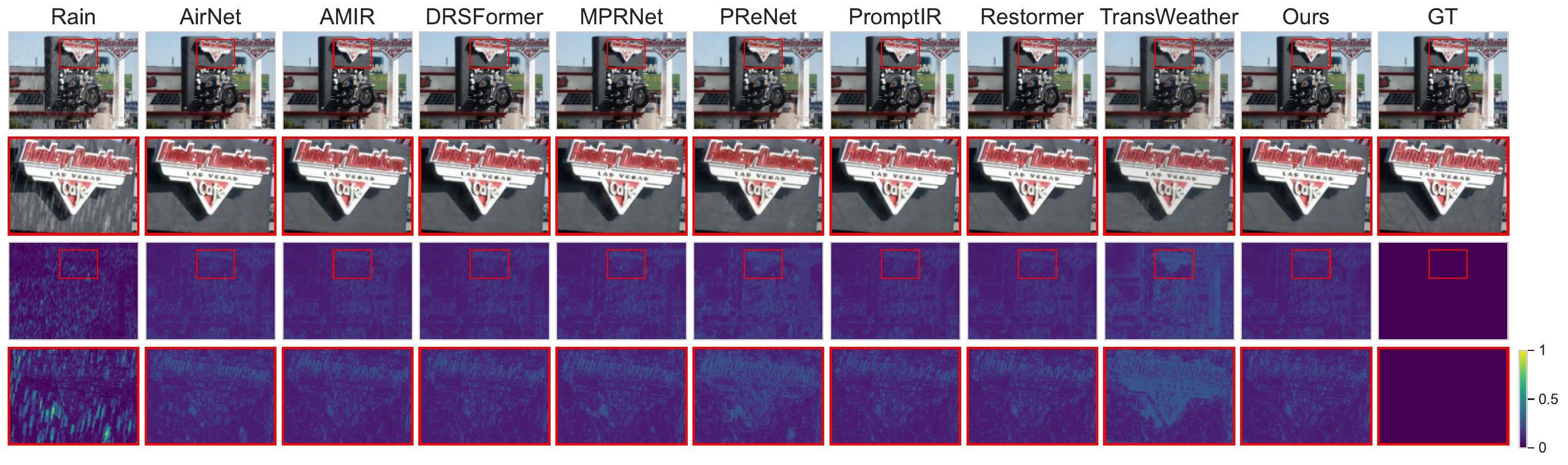}}
  \caption{Visual comparison of different deraining models. For each group of images, the first row shows the original images, the second row shows the mean absolute error of the images.}\label{fig:viz-cmp}
\end{figure}

Table~\ref{tab:cmp-size} presents a comprehensive analysis of model size, computational complexity, and inference efficiency across different deraining models. The computational complexity is measured on input images with a shape of $(1,3,256,256)$. The inference time represents the total testing time (seconds) on the Mixed dataset. Our CLIP-RPN model has 32.72 million parameters, which is comparable to state-of-the-art models like DRSformer~\cite{DRSformer} (33.63M) and PromptIR~\cite{PromptIR} (32.97M), but significantly larger than lightweight models like PReNet~\cite{PReNet} (0.17M). In terms of computational complexity measured by MACs (Multiply-Accumulate Operations), our model requires 293.53 Giga MACs, placing it among the more computationally intensive models, though still more efficient than MPRNet's~\cite{MPRNet} 548.65 Giga MACs and AirNet's~\cite{AirNet} 301.27 Giga MACs. The substantial computational requirements primarily stem from the integration of the CLIP~\cite{CLIP} model. Notably, our model has a relatively long inference time (76.90s), while TransWeather~\cite{TransWeather} achieves the fastest inference time (8.61s) with the lowest MACs (6.13G). However, combined the quantitative comparison in Table~\ref{tab:cmp-quantitative}, although our model exhibits higher computational complexity and longer inference time due to the incorporation of CLIP and multi-branch architecture, it consistently delivers superior performance across various metrics.

\subsubsection{Visual Comparison}

Fig.~\ref{fig:viz-cmp} illustrates the visual results of our model compared to other methods. Our model effectively removes rain streaks while preserving image details, outperforming other models in visual quality. The images processed by CLIP-RPN exhibit clearer details and fewer artifacts, particularly in challenging mixed rain conditions. This visual evidence underscores the quantitative improvements and validates the effectiveness of our approach in real-world applications.

\subsection{Ablation Study and Parameter Analysis}

\subsubsection{Comparison of Different Prompts}

\begin{table}[htbp]
  \centering
  \caption{Three different sets of prompts used in our experiments, generated by GPT-4~\cite{openai2024gpt4technicalreport}.}
  \small
  \resizebox{0.8\linewidth}{!}{
    \begin{tabular}{ m{2cm}<{\centering} |  m{0.7\linewidth} }
      \toprule
      Index & \hspace{4cm} Prompt Description                                                                                                                                                                                                                                                                                                                                                                                                          \\
      \midrule
      1     & \footnotesize\texttt{["Image showing a gentle drizzle with minimal water accumulation. The raindrops may be small and sparse.", "Image depicting a steady rainfall with more noticeable raindrops. Puddles may start to form, and visibility might be slightly reduced.", "Image characterized by intense rainfall, with large raindrops and significant water accumulation. Visibility is often low, and streets may appear flooded."]} \\
      \noalign{\vskip 4pt}
      \hline
      \noalign{\vskip 4pt}
      2     & \footnotesize\texttt{["This image has very sparse raindrops", "This is an image with moderately dense raindrops", "This is an image with dense raindrops"]}                                                                                                                                                                                                                                                                              \\
      \noalign{\vskip 4pt}
      \hline
      \noalign{\vskip 4pt}
      3     & \footnotesize\texttt{["This image has almost no rain effect or raindrops or any distoration.", "Rain effect: the rain looks like it was unnatural and of poor quality, adding to the image distortion."]} \bigstrut[b]                                                                                                                                                                                                                   \\
      \bottomrule
    \end{tabular}}%
  \label{tab:prompts}%
\end{table}%  

\begin{table}[htbp]
  \centering
  \small
  \caption{Quantitative results of the proposed method with different prompts.}
  \resizebox{0.75\linewidth}{!}{
    \begin{tabular}{ccccccccccc}
      \toprule
      \multirow{2}[3]{*}{Prompt}                                                                  & \multicolumn{2}{c}{Rain100L~\cite{Rain100}} & \multicolumn{2}{c}{Rain100H~\cite{Rain100}} & \multicolumn{2}{c}{Rain800~\cite{Rain800}} & \multicolumn{2}{c}{Mixed} & \multicolumn{2}{c}{Average} \bigstrut                                                                                                      \\
      \cmidrule(r){2-3} \cmidrule(r){4-5} \cmidrule(r){6-7} \cmidrule(r){8-9} \cmidrule(r){10-11} & PSNR                                        & SSIM                                        & PSNR                                       & SSIM                      & PSNR                                  & SSIM            & PSNR           & SSIM            & PSNR           & SSIM \bigstrut[t]            \\
      \midrule
      1                                                                                           & 36.19                                       & 0.9209                                      & 29.51                                      & 0.8382                    & 27.38                                 & 0.7939          & 32.13          & 0.8635          & 31.30          & 0.8541                       \\
      2                                                                                           & 36.01                                       & \textbf{0.9214}                             & 29.63                                      & 0.8421                    & 27.80                                 & 0.8058          & 32.39          & 0.8663          & 31.45          & 0.8589                       \\
      3                                                                                           & \textbf{36.21}                              & 0.9211                                      & \textbf{29.66}                             & \textbf{0.8423}           & \textbf{28.40}                        & \textbf{0.8122} & \textbf{32.42} & \textbf{0.8671} & \textbf{31.67} & \textbf{0.8607} \bigstrut[b] \\
      \bottomrule
    \end{tabular}}%
  \label{tab:cmp-prompt}%
\end{table}%

\begin{table}[htbp]
  \centering
  \vspace{0.5em}
  \caption{Analysis of rain intensity distribution across different datasets using CLIP-based prompt evaluation (Prompt 1 from Table~\ref{tab:prompts}).}
  \vspace{0.5em}
  \small
  \resizebox{\textwidth}{!}{
    \begin{tabular}{lccc}
      \toprule
      \noalign{\vskip 7pt}
      {Dataset$\backslash$Prompt}                                                                                                                                                                                                                            &
      \begin{tabular}[c]{@{}c@{}}\footnotesize\texttt{Image showing a gentle drizzle with minimal} \\ \footnotesize\texttt{water accumulation. The raindrops may be} \\ \footnotesize\texttt{small and sparse.}\end{tabular}                                 &
      \begin{tabular}[c]{@{}c@{}}\footnotesize\texttt{Image depicting a steady rainfall with more} \\ \footnotesize\texttt{noticeable raindrops. Puddles may start to} \\ \footnotesize\texttt{form, and visibility might be slightly reduced.}\end{tabular} &
      \begin{tabular}[c]{@{}c@{}}\footnotesize\texttt{Image characterized by intense rainfall, with} \\ \footnotesize\texttt{large raindrops and significant water} \\ \footnotesize\texttt{accumulation. Visibility is often low, and} \\ \footnotesize\texttt{streets may appear flooded.}\end{tabular} \bigstrut[b] \\
      \noalign{\vskip 7pt}
      \hline
      Rain800~\cite{Rain800}                                                                                                                                                                                                                                 & 33.29 & 0.86 & 65.86 \bigstrut[t]                       \\
      Rain100H~\cite{Rain100}                                                                                                                                                                                                                                & 20.89 & 0.00 & 79.11                                    \\
      Rain100L~\cite{Rain100}                                                                                                                                                                                                                                & 32.50 & 0.00 & 67.50                                    \\
      Mixed                                                                                                                                                                                                                                                  & 24.96 & 0.22 & 74.81 \bigstrut[b]                       \\
      \bottomrule
    \end{tabular}}%
  \label{tab:cls-prompt1}%
\end{table}%

\begin{table}[htbp]
  \centering
  \vspace{0.5em}
  \caption{Analysis of rain intensity distribution across different datasets using CLIP-based prompt evaluation (Prompt 2 from Table~\ref{tab:prompts}).}
  \vspace{0.5em}
  \small
  \resizebox{\textwidth}{!}{
    \begin{tabular}{lccc}
      \toprule
      \noalign{\vskip 7pt}
      {Dataset$\backslash$Prompt}                                                                                                                                                                                                        &
      \begin{tabular}[c]{@{}c@{}}\footnotesize\texttt{This image has very sparse raindrops,} \\ \footnotesize\texttt{indicating light rain conditions with} \\ \footnotesize\texttt{minimal impact on visibility.}\end{tabular}          &
      \begin{tabular}[c]{@{}c@{}}\footnotesize\texttt{This is an image with moderately dense} \\ \footnotesize\texttt{raindrops, showing noticeable rain} \\ \footnotesize\texttt{intensity and some visibility reduction.}\end{tabular} &
      \begin{tabular}[c]{@{}c@{}}\footnotesize\texttt{This is an image with dense raindrops,} \\ \footnotesize\texttt{representing heavy rain conditions with} \\ \footnotesize\texttt{significant visibility impairment.}\end{tabular} \bigstrut[b]                         \\
      \noalign{\vskip 7pt}
      \hline
      Rain800~\cite{Rain800}                                                                                                                                                                                                             & 21.71 & 4.86 & 73.43 \bigstrut[t] \\
      Rain100H~\cite{Rain100}                                                                                                                                                                                                            & 6.940 & 8.39 & 84.67              \\
      Rain100L~\cite{Rain100}                                                                                                                                                                                                            & 24.00 & 8.50 & 67.50              \\
      Mixed                                                                                                                                                                                                                              & 12.04 & 7.48 & 80.48 \bigstrut[b] \\
      \bottomrule
    \end{tabular}}%
  \label{tab:cls-prompt2}%
\end{table}%

\begin{table}[htbp]
  \centering
  \vspace{0.5em}
  \caption{Analysis of rain intensity distribution across different datasets using CLIP-based prompt evaluation (Prompt 3 from Table~\ref{tab:prompts}).}
  \vspace{0.5em}
  \small
  \resizebox{0.8\linewidth}{!}{
    \begin{tabular}{lcc}
      \toprule
      \noalign{\vskip 7pt}
      {Dataset$\backslash$Prompt}                                                                                                                                 &
      \begin{tabular}[c]{@{}c@{}}\footnotesize\texttt{This image has almost no rain effect or} \\ \footnotesize\texttt{raindrops or any distortion.}\end{tabular} &
      \begin{tabular}[c]{@{}c@{}}\footnotesize\texttt{Rain effect: the rain looks like it was} \\ \footnotesize\texttt{unnatural and of poor quality, adding to} \\ \footnotesize\texttt{the image distortion.}\end{tabular} \bigstrut[b] \\
      \noalign{\vskip 7pt}
      \hline
      Rain800~\cite{Rain800}                                                                                                                                      & 18.00 & 82.00 \bigstrut[t]                                            \\
      Rain100H~\cite{Rain100}                                                                                                                                     & 6.110 & 93.89                                                         \\
      Rain100L~\cite{Rain100}                                                                                                                                     & 29.50 & 70.50                                                         \\
      Mixed                                                                                                                                                       & 10.93 & 89.07 \bigstrut[b]                                            \\
      \bottomrule
    \end{tabular}}%
  \label{tab:cls-prompt3}%
\end{table}%

In our experiments, we utilized three distinct sets of prompts, as shown in Table~\ref{tab:prompts}. These prompts were generated by GPT-4~\cite{openai2024gpt4technicalreport} and designed to capture different rain intensities and conditions. The quantitative results of our proposed method with these different prompts are presented in Table~\ref{tab:cmp-prompt}. It is evident that on most cases, Prompt 3 consistently outperforms the other two prompts across all datasets in terms of both PSNR and SSIM metrics. Overall, Prompt 3 demonstrates the best performance, followed by Prompt 2, while Prompt 1 shows the weakest results.

To analyze the reasons for the different performances of various prompts, we conducted an analysis of rain intensity distribution across different datasets using CLIP-based prompt evaluation. Tables~\ref{tab:cls-prompt1}, \ref{tab:cls-prompt2}, and \ref{tab:cls-prompt3} present the results of this analysis for Prompts 1, 2, and 3, respectively. These results provide valuable insights into the distribution of rain intensities in the datasets and the effectiveness of the prompts in capturing these intensities. We believe that the poor performance of Prompt 1 is primarily due to insufficient activation of certain branches, as shown in Table~\ref{tab:cls-prompt1}. On the mixed dataset, only 0.22\% of images were used to train a particular branch, which may lead to inadequate network training. In contrast, Prompts 2 and 3 demonstrate relatively better performance because they have sufficient images to fully train all sub-branches. The analysis demonstrates the importance of carefully designing and selecting the prompts for our model. A well-designed prompt can significantly enhance the performance of our model, enabling it to effectively handle a wide range of rain conditions and intensities. Moreover, our model's superior performance with Prompt 3 underscores its robustness and adaptability, further highlighting the effectiveness of our approach.

\subsubsection{Ablation Study of Components}

% Table generated by Excel2LaTeX from sheet 'Sheet4'
\begin{table}[htbp]
  \centering
  \caption{Ablation study of components.}
  \small
  \resizebox{0.25\linewidth}{!}{
    \begin{tabular}{cccc}
      \hline
      \multicolumn{2}{c}{Model}                  & \multicolumn{2}{c}{Mixed Data} \bigstrut                                                 \\
      \cmidrule(r){1-2} \cmidrule(r){3-4}    RPN & MGCA                                     & PSNR           & SSIM \bigstrut               \\
      \hline
      \ding{51}                                  & \ding{51}                                & \textbf{32.42} & \textbf{0.8671} \bigstrut[t] \\
      \ding{55}                                  & \ding{51}                                & 32.31          & 0.8656                       \\
      \ding{51}                                  & \ding{55}                                & 32.11          & 0.8635                       \\
      \ding{55}                                  & \ding{55}                                & 32.08          & 0.8640 \bigstrut[b]          \\
      \hline
    \end{tabular}}%
  \label{tab:addlabel}%
\end{table}%

To understand the contributions of each component in our model, we conduct an ablation study by systematically removing or altering key components, including the CLIP-driven rain perception module (RPN) and the mask-guided cross-attention mechanism (MGCA). The results are summarized in Table~\ref{tab:cmp-loss} and reveal several important insights.

The complete model with both RPN and MGCA achieves the best performance with PSNR of 32.42 and SSIM of 0.8671, demonstrating the effectiveness of our proposed components. When removing the RPN module while keeping MGCA, the performance drops to 32.31 PSNR and 0.8656 SSIM, indicating that the rain pattern awareness and adaptive routing mechanism contribute significantly to the model's capability in handling diverse rain conditions. Similarly, removing MGCA while maintaining RPN results in a more substantial performance degradation, suggesting that the mask-guided cross-attention plays a crucial role in precisely removing rain artifacts while preserving image details. The baseline model without both components achieves the lowest performance, which is still relatively high due to the fundamental architecture of our network. These results collectively demonstrate that while both components contribute to the overall performance, the mask-guided cross-attention mechanism has a slightly more significant impact than the rain perception module, and their combination leads to the most effective deraining solution.

\subsubsection{Comparison with Commonly Used Loss}

To evaluate the effectiveness of our proposed dynamic loss scheduling (DLS) mechanism, we conduct comprehensive experiments comparing it with traditional loss functions including $l_1$, $l_2$, and Huber losses, as shown in Table~\ref{tab:cmp-loss}.

\begin{table}[htbp]
  \centering
  \caption{Comparison results between the proposed DLS and commonly used $l_1$, $l_2$, and Huber losses.}
  \small
  \resizebox{0.8\linewidth}{!}{
    \begin{tabular}{ccccccccccc}
      \hline
      \multirow{2}[4]{*}{Loss}                                                                    & \multicolumn{2}{c}{Rain100L~\cite{Rain100}} & \multicolumn{2}{c}{Rain100H~\cite{Rain100}} & \multicolumn{2}{c}{Rain800~\cite{Rain800}} & \multicolumn{2}{c}{Mixed} & \multicolumn{2}{c}{Average} \bigstrut                                                                                                      \\
      \cmidrule(r){2-3} \cmidrule(r){4-5} \cmidrule(r){6-7} \cmidrule(r){8-9} \cmidrule(r){10-11} & PSNR                                        & SSIM                                        & PSNR                                       & SSIM                      & PSNR                                  & SSIM            & PSNR           & SSIM            & PSNR           & SSIM \bigstrut               \\
      \hline
      $l_1$                                                                                       & 36.14                                       & \textbf{0.9223}                             & 29.56                                      & 0.8415                    & 28.12                                 & 0.8056          & 32.25          & 0.8641          & 31.52          & 0.8584 \bigstrut[t]          \\
      $l_2$                                                                                       & 36.19                                       & 0.9211                                      & 29.45                                      & 0.8375                    & 27.98                                 & 0.8069          & 32.18          & 0.8636          & 31.45          & 0.8573                       \\
      Huber                                                                                       & 36.16                                       & 0.9209                                      & 29.45                                      & 0.8377                    & 27.93                                 & 0.8052          & 32.19          & 0.8638          & 31.43          & 0.8569                       \\
      DLS (ours)                                                                                  & \textbf{36.21}                              & 0.9211                                      & \textbf{29.66}                             & \textbf{0.8423}           & \textbf{28.40}                        & \textbf{0.8122} & \textbf{32.42} & \textbf{0.8671} & \textbf{31.67} & \textbf{0.8607} \bigstrut[b] \\
      \hline
    \end{tabular}%
    \label{tab:cmp-loss}}%
\end{table}%

\begin{figure}[htbp]
  \centering
  \subfloat[PSNR]
  {\includegraphics[width=0.4\textwidth]{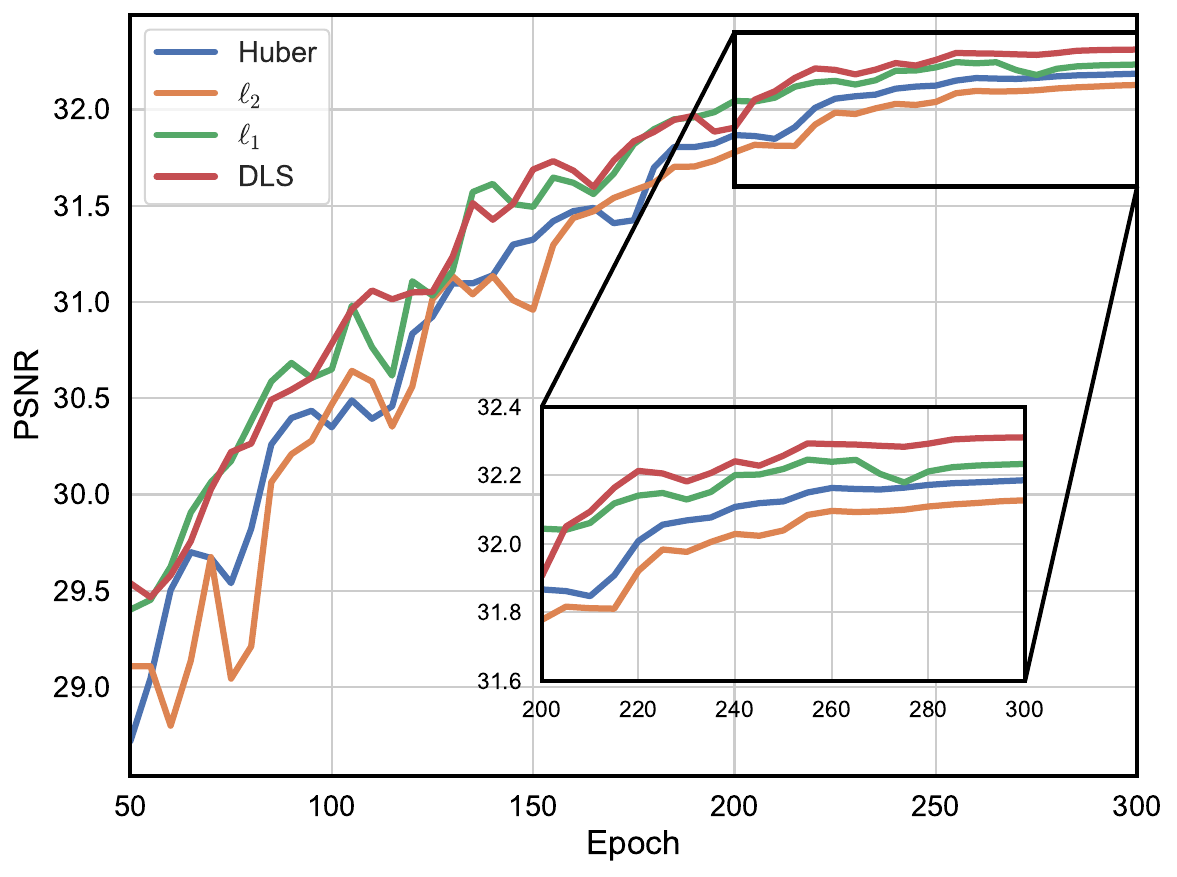}}
  \subfloat[SSIM]
  {\includegraphics[width=0.4\textwidth]{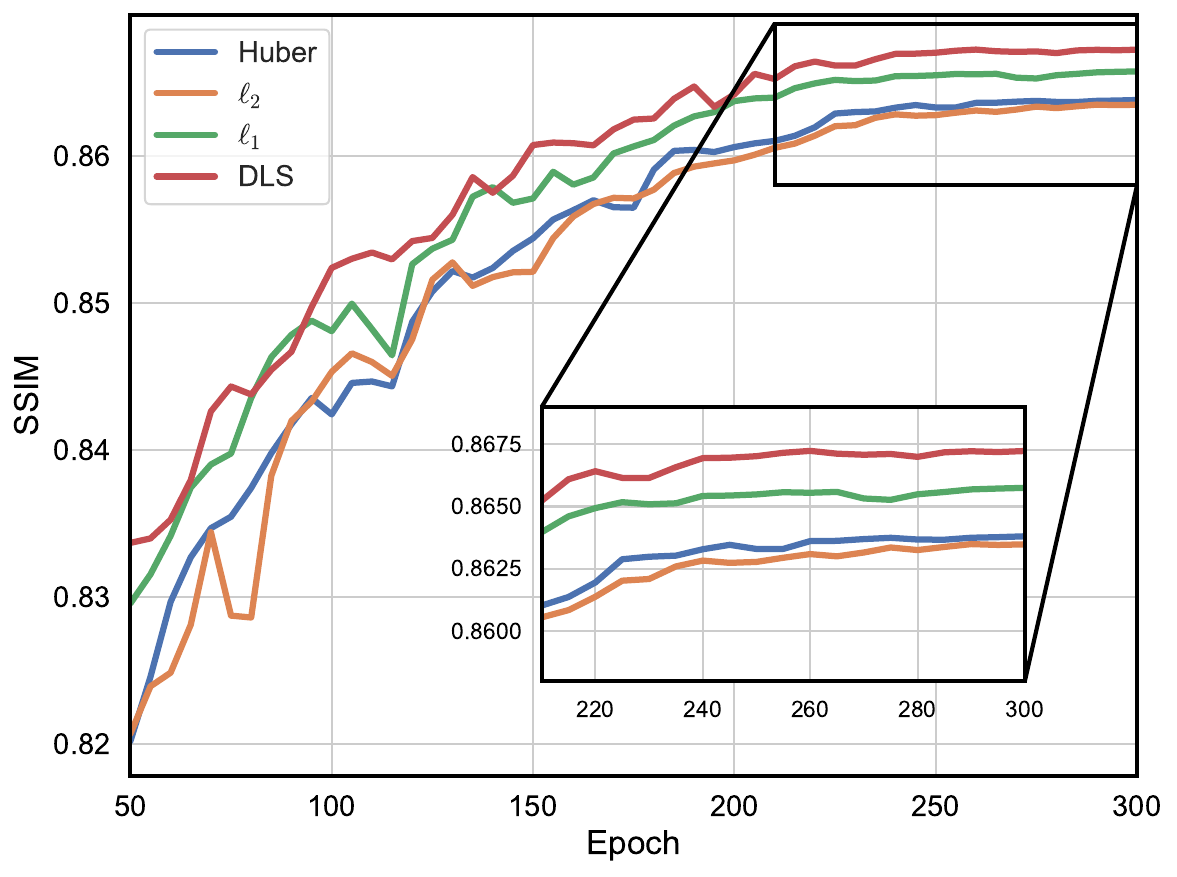}}
  \caption{Convergence curves of different losses.}\label{fig:curve-loss}
\end{figure}

DLS achieves the highest PSNR and SSIM values across all datasets, particularly excelling in the challenging Mixed dataset. The convergence curves in Fig.~\ref{fig:curve-loss} further reveal the advantages of DLS in the training process. In terms of PSNR, DLS initially performs comparably to $l_1$ loss but gradually surpasses it in later training stages, achieving better final results. For SSIM, DLS consistently outperforms other losses throughout the training process, demonstrating its ability to better preserve structural information. The superior performance of DLS can be attributed to its adaptive nature, which dynamically adjusts the loss weights based on the training progress, allowing for more effective optimization and better handling of complex rain patterns in diverse scenarios.

\subsubsection{Parameter Analysis of $\beta$ and $\eta$}

\begin{figure}[htbp]
  \centering
  \subfloat[PSNR]
  {\includegraphics[width=0.4\textwidth]{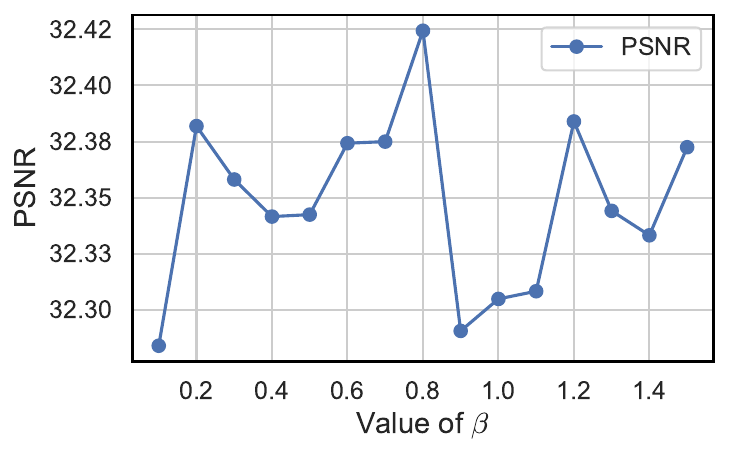}}
  \subfloat[SSIM]
  {\includegraphics[width=0.4\textwidth]{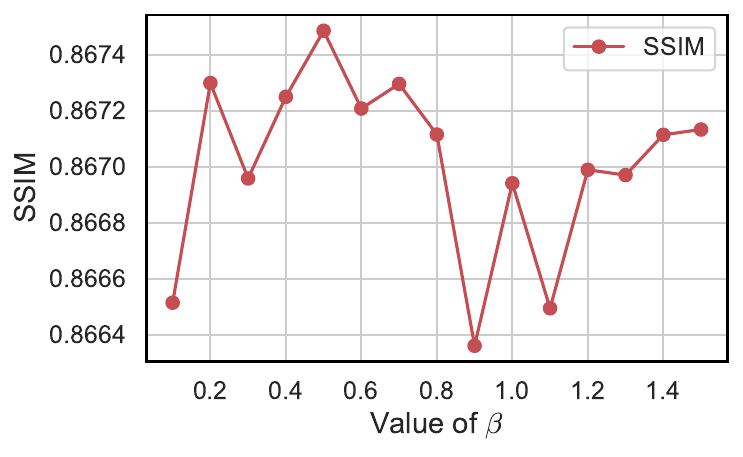}}
  \caption{Analysis of the impact of different $\beta$ values on the performance of the proposed method. $\eta$ is fixed at 2.3.}\label{fig:analysis-beta}
\end{figure}

\begin{figure}[htbp]
  \centering
  \subfloat[PSNR]
  {\includegraphics[width=0.4\textwidth]{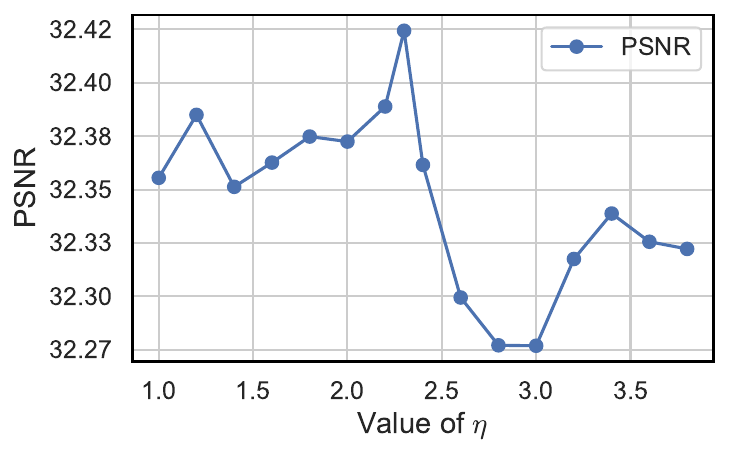}}
  \subfloat[SSIM]
  {\includegraphics[width=0.4\textwidth]{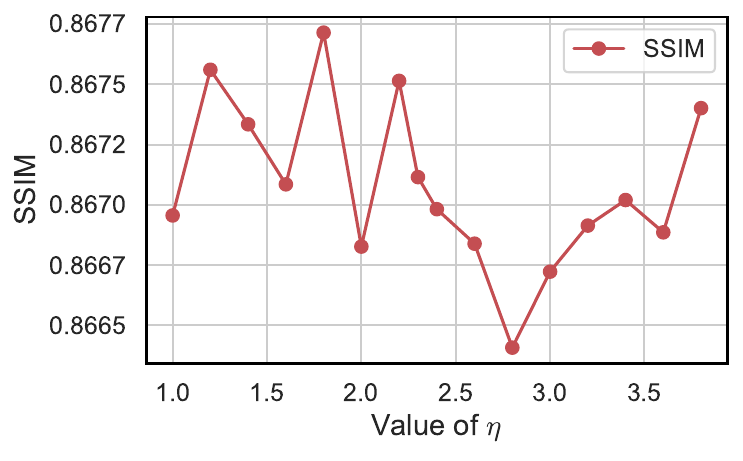}}
  \caption{Analysis of the impact of different $\eta$ values on the performance of the proposed method. $\beta$ is fixed at 0.8.}\label{fig:analysis-eta}
\end{figure}

Parameter analysis is conducted to evaluate the impact of different hyperparameters, including the $\beta$ and $\eta$ values in the dynamic loss scheduling strategy. $\beta$ influences the initial focus of the loss function (low-frequency smoothing vs. high-frequency detail preservation), while $\eta$ controls the rate at which the loss function transitions from low-frequency to high-frequency focus during training. Figs.~\ref{fig:analysis-beta} and~\ref{fig:analysis-eta} show the performance variations with different parameter settings. Overall, the model achieves optimal performance when $\beta$ and $\eta$ are set to 0.8 and 2.3, respectively. It is evident that when $\beta$ exceeds 0.8 or $\eta$ surpasses 2.3, the model performance significantly degrades. This phenomenon is reseaonable and can be explained by two factors: (1) An excessively large $\beta$ value causes the model to overemphasize pixels with high loss values, which contradicts the network's reconstruction tendency, as we analyze in Section~\ref{sec:dynamic_loss_scheduling}; (2) An overly large $\eta$ value generates excessively large gradient values, leading to training instability.

\subsubsection{Visualization of the Multi-Level Predicted Mask}

\begin{figure}[htbp]
  \centering
  \includegraphics[width=1\textwidth]{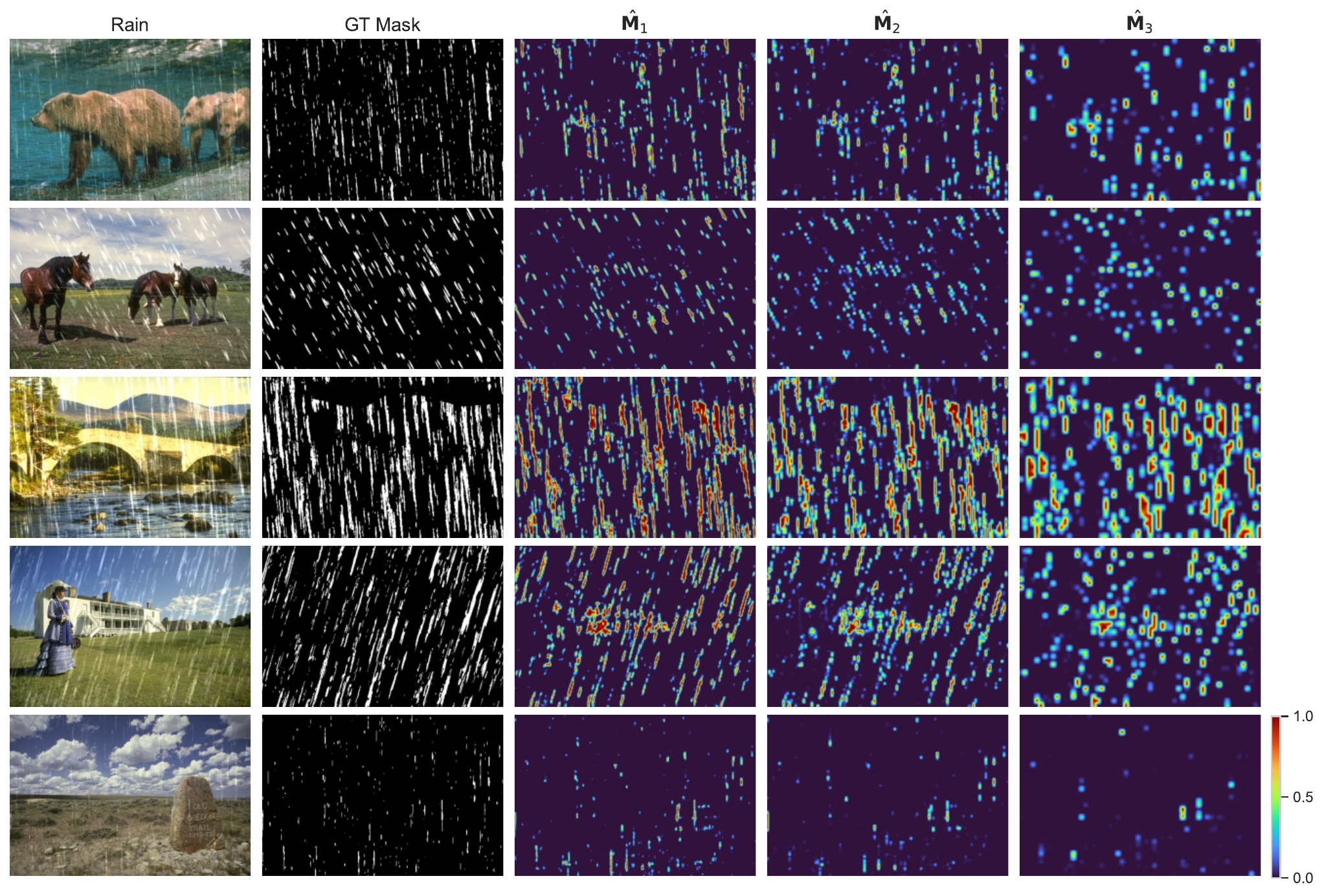}
  \caption{Visualization of the predicted mask.}\label{fig:viz-mask}
  \label{fig:viz-mask}
\end{figure}

To clearly demonstrate the effectiveness of mask-guided cross-attention, we visualize the predicted masks at each level of CLIP-RPN, as shown in \cref{fig:viz-mask}. The visualization reveals an important hierarchical pattern: shallower network layers primarily focus on capturing fine-grained details and local rain patterns, while deeper layers, benefiting from downsampling operations, progressively shift their attention towards global semantic information and rain coverage. This multi-level feature extraction mechanism enables our model to precisely localize rain regions and effectively facilitate interactions between rainy and non-rainy areas, leading to superior deraining performance. The progressive transition from local to global understanding through network depth demonstrates the model's capability to handle rain patterns at different scales, which is crucial for comprehensive rain removal in complex scenarios.

\section{Conclusion}\label{sec:conclusion}

In this paper, we proposed CLIP-RPN, a novel deraining framework that leverages CLIP's semantic alignment capabilities to perceive diverse rain patterns and adaptively route images to specialized sub-networks. By integrating a mask-guided cross-attention mechanism and a dynamic loss scheduling strategy, our model effectively handles complex rain conditions while preserving image details. Extensive experiments demonstrate that CLIP-RPN achieves state-of-the-art performance across multiple datasets, particularly excelling in challenging mixed rain scenarios. In future work, we will explore extending this framework to other weather-related image restoration tasks, such as dehazing and snow removal, and investigating more efficient architectures to reduce computational overhead.

\section*{Declaration of Competing Interest}
The authors declare that they have no known competing financial interests or personal relationships that could have appeared to influence the work reported in this paper.

\section*{CRediT Authorship Contribution Statement}

\textbf{Cong GUAN}: Methodology, Writing - original draft, Visualization. \textbf{Osamu Yoshie}: Investigation, Supervision.  

\section*{Data Availability}

Data will be made available on request.

\bibliographystyle{elsarticle-num}
\bibliography{ref.bib}

\end{document}